\documentclass[final,5p,times,twocolumn,authoryear]{elsarticle}

\usepackage{amssymb}
\usepackage{amssymb}
\usepackage{makecell}
\usepackage{booktabs}
\usepackage{graphicx}
\usepackage{multirow}
\usepackage{float}
\usepackage{subfigure}
\usepackage{color}
\usepackage{amsmath,bm}
\usepackage{changes} 

\usepackage{lineno}

\usepackage{hyperref}

\usepackage{soul} 
\usepackage{color, xcolor} 
\usepackage{color, soul}
\setulcolor{red}
\setstcolor{blue}
\sethlcolor{yellow}

\usepackage{subcaption}

\journal{ISPRS JPRS}

\begin{document}

\begin{frontmatter}



\title{UAVPairs: A Challenging Benchmark for Match Pair Retrieval of \\Large-scale UAV Images}


\author[label1,label3]{Junhuan Liu}
\author[label1,label2]{San Jiang\corref{cor1}}
\author[label3]{Wei Ge}
\author[label4]{Wei Huang}
\author[label5]{Bingxuan Guo}
\author[label1,label2]{Qingquan Li}

\affiliation[label1]{organization={Guangdong Key Laboratory of Urban Informatics, Shenzhen University},
            city={Shenzhen},
            postcode={518060}, 
            country={China}}
            
\affiliation[label2]{organization={MNR Key Laboratory for Geo-Environmental Monitoring of Great Bay Area, Shenzhen University},
            city={Shenzhen},
            postcode={518060}, 
            country={China}}
            
\affiliation[label3]{organization={School of Computer Science, China University of Geosciences},
            city={Wuhan},
            postcode={430074}, 
            country={China}}

\affiliation[label4]{organization={College of Mathematics and Computer Science, Wuhan Polytechnic University},
            city={Wuhan},
            postcode={430074}, 
            country={China}}
            
\affiliation[label5]{organization={State Key Laboratory of Information Engineering in Surveying, Mapping and Remote Sensing, Wuhan University},
            city={Wuhan},
            postcode={430074}, 
            country={China}}

\cortext[cor1]{Corresponding Author: jiangsan@szu.edu.cn}

\begin{abstract}
The primary contribution of this paper is a challenging benchmark dataset, UAVPairs, and a training pipeline designed for match pair retrieval of large-scale UAV images. First, the UAVPairs dataset, comprising 21,622 high-resolution images across 30 diverse scenes, is constructed; the 3D points and tracks generated by SfM-based 3D reconstruction are employed to define the geometric similarity of image pairs, ensuring genuinely matchable image pairs are used for training. Second, to solve the problem of expensive mining cost for global hard negative mining, a batched nontrivial sample mining strategy is proposed, leveraging the geometric similarity and multi-scene structure of the UAVPairs to generate training samples as to accelerate training. Third, recognizing the limitation of pair-based losses, the ranked list loss is designed to improve the discrimination of image retrieval models, which optimizes the global similarity structure constructed from the positive set and negative set. Finally, the effectiveness of the UAVPairs dataset and training pipeline is validated through comprehensive experiments on three distinct large-scale UAV datasets. The experiment results demonstrate that models trained with the UAVPairs dataset and the ranked list loss achieve significantly improved retrieval accuracy compared to models trained on existing datasets or with conventional losses. Furthermore, these improvements translate to enhanced view graph connectivity and higher quality of reconstructed 3D models. The models trained by the proposed approach perform more robustly compared with hand-crafted global features, particularly in challenging repetitively textured scenes and weakly textured scenes. For match pair retrieval of large-scale UAV images, the trained image retrieval models offer an effective solution. The dataset would be made publicly available at \url{https://github.com/json87/UAVPairs}.
\end{abstract}




\end{frontmatter}


\section{Introduction}
\label{sec1}


Unmanned Aerial Vehicle (UAV) has emerged as a prevalent remote sensing platform for 3D reconstruction because of its high timeliness and flexibility \citep{jiang2021unmanned}. However, constrained by sensor costs and payload limitations, most current commercial UAV platforms are not equipped with high precision and lightweight Positioning and Orientation Systems (POS). Efficient and accurate UAV image orientation constitutes a prerequisite for their widespread applications.

The incremental Structure from Motion (SfM) technique has become a prevalent solution for UAV image georeferencing as it can obtain camera poses and reconstruct 3D scenes directly from ordered or unordered overlapping images without the POS data \citep{wang2019structure}. The standard SfM workflow consists of three fundamental processing stages: (1) feature extraction, (2) feature matching, and (3) incremental reconstruction \citep{jiang2021unmanned}. Although recent advances in hardware acceleration and algorithmic optimization have substantially improved the efficiency of feature extraction, feature matching persists as the primary computational bottleneck in SfM for large-scale UAV images \citep{2016Recent,zhang2024legged}. This limitation stems principally from the high resolution and overlap inherent to UAV images.

Compared with enhancing image feature matching efficiency, employing Content-Based Image Retrieval (CBIR) to select a subset of image pairs for feature matching constitutes a more straightforward strategy \citep{jiang2021unmanned}. The common approach employs hand-crafted global features for image retrieval, such as Bag-of-Words (BoW) \citep{sivic2003video}, Vector of Locally Aggregated Descriptors (VLAD) \citep{jegou2010aggregating}, and Fisher Vector (FV) \citep{perronnin2010large}. These global features are generated by aggregating local descriptors that encode local gradient information, such as SIFT \citep{lowe2004distinctive}, SURF \citep{Herbert2008Speeded}, and ORB \citep{rublee2011orb}. Due to their inherent dependence on local gradient variations from hand-crafted local features, these global features exhibit significantly degraded discriminative performance in weak-texture scenes. In contrast, deep learning-based methods demonstrate robust discriminative capabilities in weak-texture scenes by capturing both global contextual patterns and high-level semantic information from images. In the fields of photogrammetry and computer vision, deep local features have undergone explosive development, progressing along two main directions: (1) patch description networks, including L2Net \citep{tian2017l2}, HardNet \citep{mishchuk2017working}, and GeoDesc \citep{luo2018geodesc}, and (2) joint detection-and-description networks, including SuperPoint \citep{2018SuperPoint}, D2-Net \citep{8953622}, R2D2 \citep{revaud2019r2d2}, and ASLFeat \citep{luo2020aslfeat}. Although deep local features have demonstrated superior performance over SIFT in the task of feature matching, the comparative work by \cite{liu2024matchable} reveals that they consistently underperform SIFT in match pair retrieval of UAV images. Moreover, such local feature aggregation-based methods demonstrate limited scalability, where the retrieval efficiency and accuracy deteriorate rapidly as the scale of the dataset expands.

By contrast, deep global features provide an efficient, generic, and scalable end-to-end solution. These methods derive compact global image representations from intermediate feature maps extracted by Convolutional Neural Networks (CNN), such as NetVLAD \citep{arandjelovic2016netvlad}, SpoC \citep{7410507}, GeM \citep{radenovic2018fine}, and MIRorR \citep{shen2018matchable}. With the incorporation of attention mechanisms, more advanced global feature extraction networks are proposed, including SOLAR \citep{ng2020solar}, DOLG \citep{yang2021dolg}, DALG \citep{song2022dalg}, and GLAM \citep{song2022all}. However, the existing methods are still flawed of network training, mainly in the aspects of the training dataset and the loss function. Most existing image retrieval models are trained on object or landmark retrieval datasets, such as UKBench \citep{nister2006scalable}, Holidays \citep{jegou2008hamming}, Oxford-5k \citep{philbin2007object}, Paris-6k \citep{philbin2008lost}, INSTRE \citep{wang2015instre}, GLDv1 \citep{noh2017large}, and GLDv2 \citep{weyand2020google}, which exhibit significant discrepancies with UAV images in terms of resolution and content. Moreover, image pair retrieval aims to identify potentially matching and spatially overlapping image pairs, which cannot be fine-grained defined by semantic labeling. Although the UAV datasets GL3D \citep{shen2018matchable} and LOIP \citep{HOU2023103162} are annotated with geometric similarity derived from mesh reprojection, this approach may yield image pairs with excessive viewpoint variations that surpass the matching capacity of the local feature. Crucially, match pair retrieval cannot be separated from the SfM framework, and the generated training image pairs should account for the practical matching limitations of the local feature. In addition, since the GL3D dataset only provides downsampled images and the LOIP dataset is not organized per scene, expanding these datasets as UAV scenes increase is not available.

Existing loss functions for image retrieval exhibit limitations in leveraging the fine-grained global ranking structure. Pair-based losses, such as the contrastive loss \citep{ROOPAK1993SIGNATURE} and the triplet loss \citep{schroff2015facenet}, although intuitively designed to enforce proximity between similar instances and separation between dissimilar ones, are plagued by issues of slow convergence and expensive negative or triplet mining, particularly in large-scale scenes. To improve scalability and sampling efficiency, proxy-based losses, such as Proxy-NCA \citep{movshovitz2017no}, achieve more efficient and stable training by exploiting class proxy vectors, but they overlook fine-grained intra-class information, limiting the ability of fine-grained ranking. More recently, classification-based losses adapted for image retrieval, such as ArcFace \citep{8953658} and CosFace \citep{wang2018cosface}, incorporate angular or cosine margins into the Softmax loss to enhance inter-class separation and intra-class compactness. However, these methods require images to correspond to a semantic category, which is contrary to the match pair retrieval task that focuses on geometric similarity rather than semantic similarity.

To address the mentioned issues regarding network training of deep global features, this paper makes three main contributions: (1) the UAVPairs dataset for match pair retrieval of large-scale UAV Images is constructed. To obtain genuine matching correlations of image pairs, SfM-based 3D reconstruction is performed for each scene and the geometric similarity is defined with the number of common 3D points. Image pairs produced in this manner are guaranteed to be matchable, and subsequent 3D reconstruction serves as a filtering step to eliminate mismatched image pairs. (2) Since proxy-based losses overlook fine-grained intra-class information and classification-based losses are unsuitable for match pair retrieval, pair-based losses are still used. However, given the expensive global hard negative mining in \cite{radenovic2018fine}, a batched nontrivial sample mining strategy is proposed to decrease the sample mining cost and accelerate the network training. (3) Recognizing that the contrastive loss and the triplet loss focus solely on the local similarity structure of a pair or triplet, the ranked list loss that leverages the global similarity structure of the query is proposed to enhance the discrimination of deep global features. By using real datasets, the proposed solution is extensively evaluated in image retrieval and SfM reconstruction.

This paper is organized as follows. Section \ref{sec2} introduces the UAVPairs benchmark dataset, along with a comparison to other image retrieval datasets. Section \ref{sec3} details the proposed image retreival method, and Section \ref{sec4} describes the conducted experiments, including test datasets, evaluation metrics, and presents the results for match pair retrieval and SfM-based reconstruction. Finally, Section \ref{sec5} presents the conclusion.

\section{UAVPairs: a scalable dataset for match pair retrieval of UAV images}
\label{sec2}
\subsection{The UAVPairs benchmark }
\label{sec2.1}

\begin{table*}[!ht]
    \centering
    \caption{The statistics and various scene characteristics of the UAVPairs dataset.}
    \label{tab:Table 1}   
\begin{tabular}{llrl} 
\toprule 
        Scene Categories & Scenes & Images & Scene Characteristics \\        
\midrule 
        Rural farmland & 9 & 4,709 & Coverage of sparse buildings and farmland \\
        Urban blocks & 6 & 6,003 & Buildings with obstructions and shadows \\
        River corridors & 2 & 604 & Covered with water, weak texture area \\
        Mountain areas & 6 & 3,455 & A large area of vegetation, undulating terrain \\
        Groups of buildings & 5 & 2,953 & Dense buildings, repetitive structures \\
        Hybrid scenes & 2 & 4,502 & Large area with multiple land cover categories \\        
\bottomrule 
\end{tabular}
\end{table*}

\begin{figure*}[!ht]
\centering
\subfigure[ rural farmland]{
\includegraphics[width=0.25\linewidth]{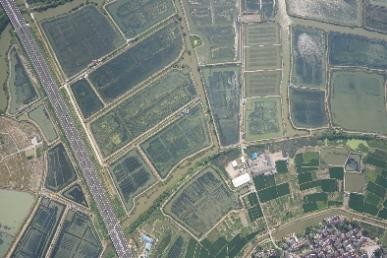}
}
\subfigure[urban blocks]{
\includegraphics[width=0.25\linewidth]{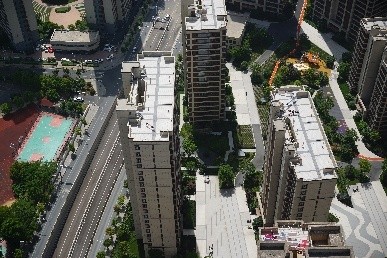}
}
\subfigure[river corridors]{
\includegraphics[width=0.25\linewidth]{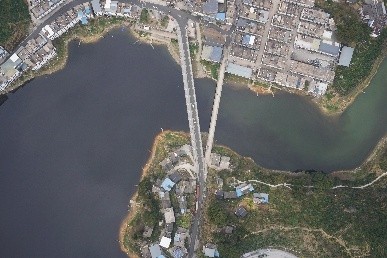}
}

\subfigure[mountain areas]{
\includegraphics[width=0.25\linewidth]{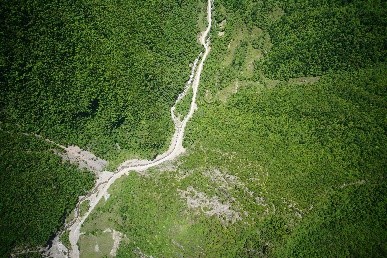}
}
\subfigure[buildings groups]{
\includegraphics[width=0.25\linewidth]{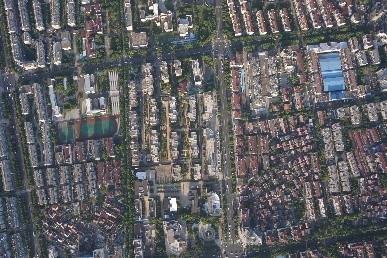}
}
\subfigure[hybrid scenes]{
\includegraphics[width=0.25\linewidth]{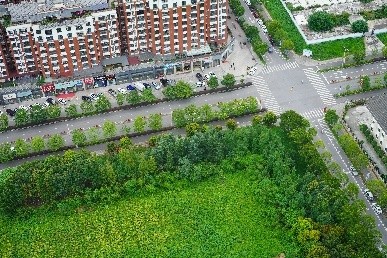}
}
\caption{UAV images of various scenes}
\label{fig:figure 1}
\end{figure*}

Most current image retrieval models are trained for instance retrieval or landmark retrieval tasks, where the training images significantly differ from UAV aerial images in terms of the captured viewpoints, observation scales, target details, and background contents. To construct the UAVPairs dataset, 21,622 high-resolution UAV images captured from multiple scales and perspectives across 30 distinct scenes are collected. Each scene contains 100 to 4,000 images with substantial geometric overlap. The ground cover categories encompass rural farmland, urban blocks, river corridors, mountainous regions, architectural complexes, and mixed scenes. The dataset statistics and scene characteristics are detailed in Table \ref{tab:Table 1}, with representative UAV image samples from each category illustrated in Figure \ref{fig:figure 1}.

The existing instance or landmark retrieval datasets, such as Oxford5k, Paris6k, GLDv1, and GLDv2, typically contain instance-level or landmark-level semantic annotations corresponding to salient individual objects in the image. In contrast, a UAV image contains plenty of various objects, thus is inadvisable to determine the similarity of images by object categories. Match pair retrieval aims to identify image pairs with high spatial overlap, emphasizing geometric context and spatial relationships rather than specific objects. As SfM-based 3D reconstruction can effectively filter out nearly all mismatching images while the generated 3D point tracks accurately characterize the overlapping relationships between image pairs, we employ SfM-based 3D reconstruction for the automatic annotation of the UAVPairs dataset.

\subsection{Auto-annotation with SfM-based 3D reconstruction}
\label{sec2.2}


\begin{figure*}[!ht]
\centering
\subfigure[rural farmland]{
\includegraphics[height=0.2\linewidth]{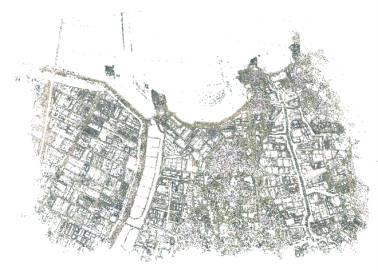}
}
\subfigure[urban blocks]{
\includegraphics[height=0.2\linewidth]{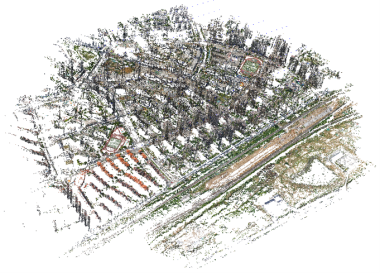}
}
\subfigure[river corridors]{
\includegraphics[height=0.2\linewidth]{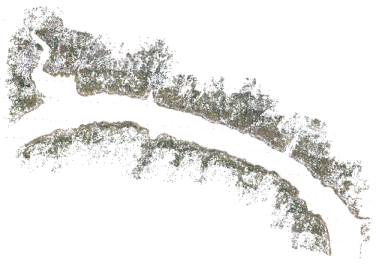}
}
\subfigure[mountain areas]{
\includegraphics[height=0.2\linewidth]{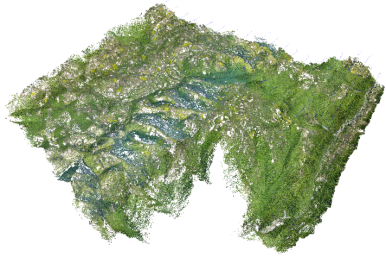}
}
\subfigure[building groups]{
\includegraphics[height=0.2\linewidth]{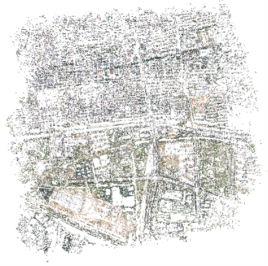}
}
\subfigure[hybrid scenes]{
\includegraphics[height=0.2\linewidth]{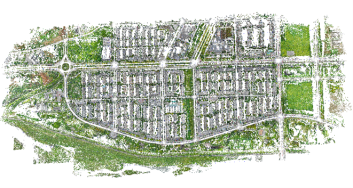}
}
\caption{UAV imagThe reconstructed 3D models of various scenes}
\label{fig:figure 3}
\end{figure*}

Since the UAVPairs dataset consists of numerous large-scale UAV images, the parallel SfM pipeline proposed by \cite{jiang2022parallel} is exploited to enhance the completeness and efficiency of automatic annotation \footnote{\url{https://github.com/json87/ParallelSfM}}. 
The pipeline utilizes image pairs retrieved via BoW to guide feature matching and facilitate 3D reconstruction. To reconstruct large-scale UAV images, we employ a divide-and-conquer strategy within the SfM framework. This methodology segments the complete scene into small-size clusters that permit both rapid and precise reconstruction. The workflow initiates with the construction of a view graph $G=(V,E)$, where vertices $V=\{v_i\}$ represent individual images and edges $E=\{e_{ij}\}$ correspond to matched image pairs $\{p_{ij}\}$. Each edge carries a significant metric ${w_{ij}}$ computed as:

\begin{equation}
    w_{ij}=R_{ew}\times w_{inlier}+(1-R_{ew})\times w_{overlap} 
    \label{eq1}
\end{equation}
where  $R_{ew}$ denotes the weighting coefficient, $w_{inlier}$ reflects the quantity of matches, and  $w_{overlap}$ represents the spatial distribution of matches. Specifically:

\begin{equation}
    w_{inlier}=\frac{log_{}N_{inlier}}{log_{}N_{maxinlier}} 
    \label{eq2}
\end{equation}

\begin{equation}
    w_{overlap}=\frac{CH_i+CH_j}{A_i+A_j} 
    \label{eq3}
\end{equation}
with  $N_{inlier}$ being the number of inliers for pair $p_{ij}$, $N_{max inlier}$ the number of maximum observed inliers among all matching pairs, $CH_i$ the convex hull area of the matching pair $p_{ij}$ on image$i$ , and $A_i$ the planar area of image $i$ . The weighted view graph is subsequently partitioned into sub-clusters using the Normalized Cut (NC) algorithm \citep{shi2000normalized} based on edge weights, resulting in strongly intra-connected and weak inter-connected sub-clusters. Subsequently, incremental SfM \citep{schonberger2016structure} is performed in parallel for each sub-cluster to generate sub-models, which are then merged sequentially according to the number of shared 3D points between models. Following the iterative merging of all sub-models, a global bundle adjustment is performed to obtain the final reconstructed 3D model. Figure \ref{fig:figure 3} presents the reconstructed 3D models of various scenes.

Suppose the dataset consists of N scenes, denoted as $S_1$…$S_i$…$S_n$, a scene $S_i$ corresponds to a 3D model $M_i=(I_i,P_i)$, where $I_i$ is the set of registered images and $P_i$ is the set of reconstructed 3D points. For a 3D point $P_i^j$, its track $T_i^j$ represents all the images associated with it. The number of common 3D points between two images can be obtained by traversing all the 3D points. Generally, image pairs with more matches have more common 3D points, while a few common 3D points also imply few matches. Therefore, the geometric similarity $GS(a,b)$ of image pairs $a$ and $b$ is defined by leveraging the number of common 3D points, as shown in Formula \ref{eq4}, where  $P_i(a)$  denotes the 3D points observed by image $a$, $P_i(b)$ represents those observed by image $b$, and $P_i(a) \cap P_i(b)$ indicates their common 3D points.
\begin{equation}    
GS(a,b)=|P_i(a) \cap P_i(b)| 
\label{eq4}
\end{equation}

\subsection{Compared with other image retrieval datasets}
\label{sec2.3}

The comparison of varying datasets for image retrieval is listed in Table \ref{tab:Table 2}, and the details are presented as follows.

\begin{table*}[!h]    
\centering    
\caption{Comparison of the UAVPairs dataset with other image retrieval datasets (* indicates that the GL3D dataset contains both UAV and non-UAV images)}    
\label{tab:Table 2}   
\begin{tabular}{llllll} 
\toprule 
        Dataset & Images & Annotation & UAV image & High Resolution & Scene Split  \\        
\midrule 
        Pittsburgh & 250K & GPS tag & × & × & × \\
        Flickr & 7.4M & 3D points & × & × & $\checkmark$ \\
        GL3D & 90K & Mesh & * & × & $\checkmark$ \\
        LOIP-PG & 10.1K & Mesh & $\checkmark$ & $\checkmark$ & $\checkmark$ \\
        UAVPairs & 21.6K & 3D points & $\checkmark$ & $\checkmark$ & $\checkmark$ \\     
\bottomrule 
\end{tabular}
\end{table*}

\textbf{Pittsburgh dataset} is a visual place recognition dataset containing 250K perspective images with 640 × 480 pixels generated from 10K Google Street View panoramas of the Pittsburgh region \citep{arandjelovic2016netvlad}. These street-view images are significantly different from UAV images in terms of imaging perspective, resolution, image content, etc. Each perspective image in the Pittsburgh dataset is associated with the GPS position of the source panorama, but two geographically close perspective images may not overlap spatially on account of different orientations or occlusions. Therefore, GPS tags are used as weakly supervised information, and the positive sample is the image with the closest distance in CNN descriptor space to the query among multiple images with close GPS. This enables the network to optimize only the optimum positive samples which results in small loss and neglects hard positive samples with weak geometric overlap but still matching.

\textbf{Flickr dataset} contains 7.4 million images downloaded from Flickr, photographing scenes primarily of famous landmarks, cities, countries, and architectural sites \citep{radenovic2018fine}. Although there are variations in spatial resolution and viewpoints between the Flickr dataset and the UAV datasets, its annotation generation method matches the UAV datasets well. For annotation generation, the clustering algorithm is first adopted to divide the scenes, and then the 3D models are reconstructed based on the clustered images by the state-of-the-art SfM. The number of co-view 3D points can serve as annotations to describe the geometric overlap between image pairs.

\textbf{GL3D dataset} is created for large-scale match pair retrieval and contains 90,590 images of 378 scenes, with both UAV scenes and non-UAV scenes \citep{shen2018matchable}. The pipeline of the annotation generation integrates dense reconstruction and surface reconstruction in addition to SfM. As the outcomes of SfM rely on local feature matching and some overlapping images are treated as unmatched due to large viewpoint differences failing feature matching, the mesh re-projection is leveraged in GL3D to accurately define the overlap region between image pairs. However, in the SfM workflow, the output of match pair retrieval serves as the input for feature matching, and the unmatched but overlapping pairs reserved in the retrieval phase will still be removed after feature matching. There is no benefit to selecting these image pairs for training. In addition, as the GL3D dataset only provides downsampled images which are unable to accomplish 3D reconstruction, it is impossible to enrich the dataset as more images from different land cover scenes become available.

\textbf{LOIP-PG dataset} is comprised of 10,097 high-resolution photogrammetric images of multiple areas ranging from forests, villages, scenic spots, cultural relics, etc \citep{HOU2023103162}. This dataset utilizes the mesh re-projection as in the 
GL3D dataset to define the similarity of image pairs to guide sample generation, as well as result in the same issue. Empirically, hard negative sample mining should be performed in scenes other than the query image scene. However, the LOIP-PG dataset does not organize the images per scene, rendering it impossible to select the hard negative sample iteratively. Moreover, this keeps this dataset from expanding with richer image sources unless additional clustering procedures to split the scenes.

Among these datasets that define matching image pairs by geometric information, the Pittsburgh dataset and the Flickr dataset are both low-resolution non-UAV images. Although the GL3D dataset and the LOIP-PG dataset contain a large amount of UAV images from multi-scenes, their sample generation methods are complex and will select overlapping pairs with no benefit for feature matching. Due to the GL3D dataset not providing the raw images and the LOIP-PG dataset not classifying the images by scenes, they are unable to be expanded as the UAV images increase to accommodate more varied scenes. Therefore, we create the UAVPairs benchmark dataset automatically annotated leveraging the outcomes of SfM, and provide raw images organized by scenes and the state-of-the-art parallel SfM pipeline to enable the dataset scalable.

\section{Methods}
\label{sec3}

\begin{figure*}[!ht]    
\centering    
\includegraphics[width=0.8\linewidth]{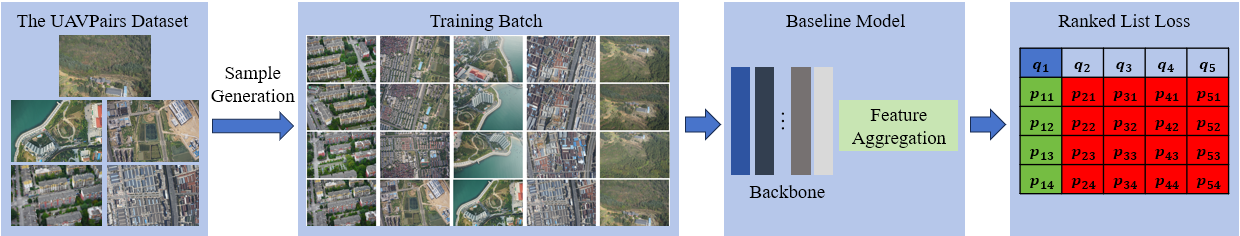}    
\caption{The overview workflow of the training pipeline.}    
\label{fig:figure 4}
\end{figure*}

To facilitate model training with the UAVPairs dataset, a comprehensive training pipeline is proposed in this study as shown in Figure \ref{fig:figure 4}. The pipeline consists of three key components: (1) A batched nontrivial sample mining strategy that generates training samples by leveraging both the geometric similarity and multi-scene structure of the UAVPairs dataset, while exclusively considering the nontrivial sample with non-zero loss during optimization; (2) A ranked list loss that operates on global similarity structures composed of matching images of the query and non-matching images from other scenes; (3) End-to-end training of baseline models consisting of a backbone and a feature aggregation layer is performed with the generated training samples and the ranked list loss. The following is the introduction of each component.

\subsection{Batched nontrivial sample generation}
\label{sec3.1}

Contrastive learning is commonly applied to image retrieval, where the learning goal is to bring the positive samples close and push the negative samples far away from the query sample. This raises the issue of how to select the positive samples $M(q)$ and negative samples $N(q)$ for the query sample $q$.

\textbf{Positive sample}. The previous method \citep{arandjelovic2016netvlad} first determines a candidate set of positive samples $CM(q)$ by weakly supervised information such as GPS due to the lack of precise geometric information to indicate the overlap of images. Then, the one in the candidate set with the minimum CNN descriptor distance to the query sample is determined as the positive sample, as in Formula \ref{eq5}. This results in selecting only the most easily optimized positive samples, so that the network will not learn much from the positive samples. In addition, the match pair retrieval requires finding all the images that match with the query image in a scene as much as possible, the hard positive samples should be brought in.
\begin{equation}    
M(q)= \underset{m \in \mathrm{CM}(q)}{\mathrm{argmin}}||f_q-f_m||
\label{eq5}
\end{equation}

The geometric similarity defined with the 3D models of the UAVPairs dataset provides a solution, which can not only obtain all matching images with the query but also fine-grained differentiate the matching degree of the matching images. Suppose the query $q$ comes from the scene $S_i$, a positive sample set that consists of images with geometric similarity greater than a threshold $\epsilon$ is constructed first. The positive samples are then randomly selected from the set as in Formula \ref{eq6}.
\begin{equation}    
M(q)=\mathrm{random}\{t\in I_i: GS(t,q)>\varepsilon\}
\label{eq6}
\end{equation}

\begin{figure*}[!ht]    
\centering    
\includegraphics[width=0.8\linewidth]{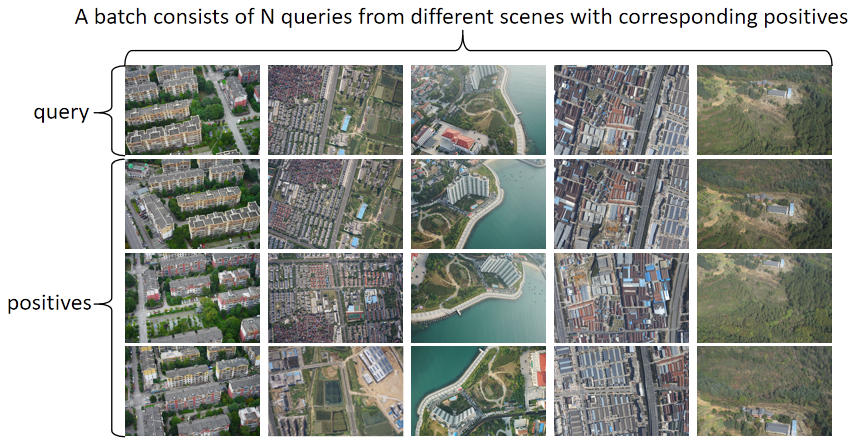}
\caption{A training batch example generated based on the batched nontrivial sample mining strategy}    
\label{fig:figure 5}
\end{figure*}

\textbf{Negative sample}. Leveraging the multi-scene structure of the UAVPairs dataset, the negative samples are selected from scenes other than the query image scene. \cite{radenovic2018fine} employs a global hard negative mining approach, which initially utilizes the pretrained network to extract the descriptors of all the images in the dataset, then selects the sample with the minimum descriptor distance to the query $q$ of the scene $S_i$ from other scenes as the hard negative sample, presented in Formula \ref{eq7}. As trained with a fixed number of steps, the image descriptors are updated and this mining step is performed again. This approach requires extracting image descriptors iteratively, causing expensive resource and time consumption.
\begin{equation}    
N(q)=\mathrm{random}\{\underset{n \in \mathrm{I_k}}{\mathrm{argmin}}||f_q-f_n||\mid k!=i\}\
\label{eq7}
\end{equation}

To improve the training efficiency, a batched nontrivial sample mining strategy without extracting image descriptors is proposed. Firstly, we randomly select $B$ scenes and a query image from each scene, denoted as $q_i$, where $i=1, \ldots,B$ . Then $M$ positive samples for each query are selected relying on the geometric similarity, denoted as $p_i^j$, where $j=1, \ldots,M$. A training batch consists of the $B\times(M+1)$  samples, the negative samples of $q_i$ are the positive samples of other queries, as in Formula \ref{eq8}. Since this procedure ignores the image descriptors, there will be many trivial samples with zero loss as training continues. These trivial samples attenuate the contribution of non-trivial samples in gradient averaging, so they are eliminated in the loss calculation. We refer to this method as batched nontrivial sample mining, a mining example is illustrated in Figure \ref{fig:figure 5}.

\begin{equation}    
N(q_i)={\{p_k^j\mid k!=i\}}
\label{eq8}
\end{equation}

\subsection{Ranked list loss}
\label{3.2}

Triplet loss is commonly used to learn discriminative feature embedding. It is defined as in Formula \ref{eq9}, where $[*]_+$ denotes the function $max(0,*)$, $D(A,P)$ denotes the descriptor distance between the anchor $A$. and the positive $P$ , $D(A,N)$ denotes the descriptor distance between the anchor $A$ and the negative $N$, and $m$ is a predefined margin used to control the minimum distance difference between positive and negative.

\begin{equation}    
L_{triplet}=[D(A,P)-D(A,N)+m]_+
\label{eq9}
\end{equation}

\begin{figure}[!h]    
\centering    
\includegraphics[width=0.7\linewidth]{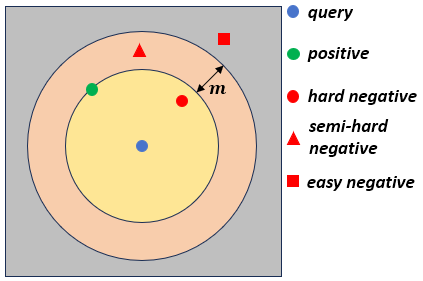}    
\caption{Three possibilities for the negative of a triplet.}    
\label{fig:figure 6}
\end{figure}

\begin{figure}[!h]
\centering
\subfigure[Triplet loss]{
\includegraphics[height=0.3\linewidth]{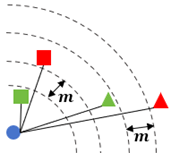}
}
\subfigure[Ranked list loss]{
\includegraphics[height=0.3\linewidth]{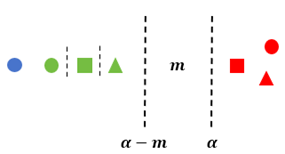}
}
\caption{Comparison of the triplet loss and the ranked list loss. (blue indicates the query sample, green indicates the positive sample, red indicates the negative sample and the same shape indicates that it is from the same triplet)}
\label{fig:figure 7}
\end{figure}

For a selected query and positive, there are three possibilities for the negative, i.e., easy negative, semi-hard negative, and hard negative, as shown in Figure \ref{fig:figure 6}. The easy triplet already satisfies the ranking and does not contribute to model optimization. If there are too many easy triplets in the training, the model cannot adequately learn discriminative feature representation. Although the hard triplet has a large loss, excessive use may lead to the model getting stuck in a local optimum or cause training oscillation. Thus, the semi-hard triplet that violates the ranking but is relatively stable is required to be mined for training. As the scale of the dataset grows, the number of triplets increases exponentially, and the cost of mining semi-hard triples also increases. In addition, triplet loss only optimizes the local ranking within an individual triplet, which may lead to the issue that the samples are ranked correctly intra-triplet but incorrectly inter-triplet, as shown in Figure \ref{fig:figure 7}(a).

To overcome the drawbacks, the ranked list loss is proposed, which directly optimizes the global ranked list consisting of the positive set and the negative set, instead of optimizing the ranking of each positive and negative pair individually, as shown in Figure \ref{fig:figure 7}(b). By optimizing the global similarity structure, the model can not only ensure correct inter-triplet ranking but also capture fine-grained differences among positives so that the more likely matched positive gets a higher similarity score. The ranked list loss consists of two terms, as shown in Formula \ref{eq10},  where $L_1$ denotes the optimization of the positive and negative set, as shown in Formula \ref{eq11}, $P$ and $N$ denote the positive set and the negative set, respectively, $\alpha$ denotes the margin of the negative and the query, and $m$ denotes the margin of the positive and the negative. The positive set is constrained to be inside a hypersphere with radius $\alpha-m$ by optimizing $L_1$, while the negative set will be pushed outside the hypersphere with radius $\alpha$.  $L_2$ denotes the optimization of the internal ranking of the positive set $P$ that is ranked in descending similarity order, as shown in Formula \ref{eq12}. The difference in matching images can be better distinguished by optimizing the $L_2$.

\begin{equation}    
L_{ranked\ list}=L_1(q,P,N)+L_2(q,P)
\label{eq10}
\end{equation}

\begin{equation}    
L_1(q,P,N) = \frac{1}{|P| + |N|} \left( \sum_{n = 1}^{N} [\alpha - D(A,n)]_+ + \sum_{p = 1}^{P} [D(A,p) - \alpha + m]_+ \right)
\label{eq11}
\end{equation}

\begin{equation}    
L_2(q,P) = \frac{1}{|P|} \left( \sum_{p = 1}^{P} [D(A,p) - D(A,p + 1)]_+ \right)
\label{eq12}
\end{equation}

In combination with the batch sample generation in Section \ref{sec3.1}, the ranked list loss of a batch $L_{batch-rll}$ is defined as in Formula \ref{eq13}, where $|B|$ denotes the number of queries in the batch, $q_i$ denote a query, $P_i$ denote the positive samples of query $q_i$ and sorted by geometric similarity. The negative samples consist of positive samples from other queries.

\begin{equation}    
L_{\text{batch - rll}} = \frac{1}{|B|} \sum_{i = 1}^{|B|} L_1 \left( q_i, P_i, \bigcup_{j = 1}^{|B|} P_j \right) + L_2(q_i, P_i), j \neq i
\label{eq13}
\end{equation}

\subsection{Baseline Models}
\label{sec3.3}

With the generated training samples and the ranked list loss, it is available to train the baseline models. The baseline models for image retrieval typically consist of a fully convolutional feature extraction backbone and a feature aggregation layer that aggregates deep feature map into compact global descriptor. Three baseline models are selected for training including NetVLAD \citep{arandjelovic2016netvlad}, GeM \citep{radenovic2018fine}, and MIRorR \citep{shen2018matchable}, as NetVLAD incorporates a powerful feature aggregation layer whereas GeM and MIRorR are similar to our training method considering geometric similarity. For NetVLAD, the $D\times H\times W$ feature map is represented as $N=H \times W$ local features $x_i$ of $D$ dimension where $i={1,\ldots,N}$. These local features are aggregated into a global feature through the feature aggregation layer NetVLAD. For GeM and MIRorR, the extracted $D \times H\times W$ feature map is represented as $H\times W$ feature map $x_k$ with index $k \in \{1, \ldots, D\}$, the feature pooling is performed on each feature map $x_k$ and the pooled feature $f_k$ of each channel are concatenated yielding the global descriptor, where the pooling layers are the GeM (Generalized Mean pooling) pooling and the max pooling, respectively. Feature aggregation is conducted as follows:

\textbf{NetVLAD:} NetVLAD is a differentiable VLAD layer, which is designed to aggregate local features extracted by FCN (Fully Convolutional Network) and support end-to-end training. The original VLAD is illustrated in Formula \ref{eq14}, where $c_k$ denotes the $k$-th word of the pre-trained codebook $C={c_1,\ldots,c_k,\ldots,c_K}$ and $x_i$ denotes the $i$-th local feature, the assignment function $a_{(i,k)=1}$ when the nearest word in the codebook to $x_i$ is $c_k$ and  $a_{(i,k)=0}$ otherwise. The VLAD layer cannot be directly embedded into CNN due to the hard assignment function  $a_{(i,k)}$ is not differentiable.

\begin{equation}    
\nu_k = \sum_{i = 1}^{n} a_{i,k}(x_i - c_k)
\label{eq14}
\end{equation}

A soft assignment function is formed via the distance between the clustering centers and the local features in place of the hard assignment function to make the VLAD layer differentiable. As shown in Formula \ref{eq15}, $\alpha$ controls the decay of the assignment value with the distance, a larger $\alpha$ means a harder assignment.

\begin{equation}    
\bar{a}_k(x_i) = \frac{e^{-\alpha \lVert x_i - c_k \rVert^2}}{\sum_{k'} e^{-\alpha \lVert x_i - c_{k'} \rVert^2}}
\label{eq15}
\end{equation}

Expanding Formula \ref{eq15} and canceling $e^{-\alpha \lVert x_i - c_k \rVert^2}$ in the numerator and denominator yields Formula \ref{eq16}, where $w_k=2ac_k$, $b_k=-\alpha \lVert c_k \rVert^2$. The parameters $w_k$, $b_k$ and $c_k$ in NetVLAD are trainable, implying that the clustering centers and the assignment function are learnable. However, in practical experiments, a simpler assignment function is adopted as it converges faster, i.e., $b_k$ is fixed to 0 and $w_k$ is initialized to $\alpha\frac{c_k}{\lVert c_k \rVert}$ in Formula \ref{eq16}. The premise is that the extracted local features are L2-normalized.

\begin{equation}    
\bar{a}_k(x_i) = \frac{e^{w_k^T x_i + b_k}}{\sum_{k'} e^{w_{k'}^T x_i + b_{k'}}}
\label{eq16}
\end{equation}

\textbf{GeM:} GeM pooling generalizes max pooling and average pooling via a learnable parameter $p$ of each feature map $X_x$ as in Formula \ref{eq17}. The parameters are end-to-end optimized with the backbone to automatically find the optimal pooling strategy for the task. Since GeM pooling can flexibly focus on global or local features, it exhibits enhanced performance compared to standard non-trained pooling layers in image retrieval.

\begin{equation}    
f_k = \left( \frac{1}{|X_k|} \sum_{x \in X_k} x^{p_k} \right)^{\frac{1}{p_k}}
\label{eq17}
\end{equation}

\textbf{MIRorR:} The models of MIRorR are trained with the GL3D dataset, which is most relevant to the match pair retrieval task as it not only contains a large number of UAV image scenes but is also annotated with the geometric similarity defined by the mesh model. Since the good translation invariance, max pooling is used for feature aggregation of MIRorR. As shown in Formula \ref{eq18}, max pooling retains the most salient feature in each feature map$X_k$.

\begin{equation}    
f_k = \max_{x \in X_k} x
\label{eq18}
\end{equation}

\section{Experiments and results}
\label{sec4}

In this section, three UAV datasets are used to evaluate the performance. First, the evaluation of match pair retrieval is performed to verify the effectiveness of the UAVPairs dataset and the ranked list loss. Second, the outcome of match pair retrieval is leveraged to guide the SfM-based 3D reconstruction, as well as to prove the performance improvement in terms of both view graph construction and 3D reconstruction. Finally, the deep global feature with the best performance is compared with the hand-crafted global features in terms of match pair retrieval and SfM-based 3D reconstruction.

\subsection{Test datasets and evaluation metrics}
\label{sec4.1}

\subsubsection{Test datasets}
\label{sec4.1.1}

\begin{figure}[!ht] 
\centering
\subfigure{
\begin{minipage}[a]{0.30\linewidth}
\centering
\includegraphics[width=1\linewidth]{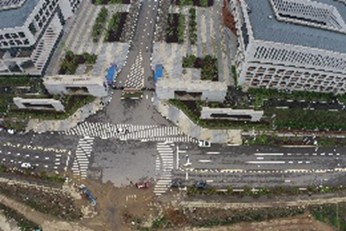} 
\end{minipage}
}
\subfigure{
\begin{minipage}[a]{0.30\linewidth}
\centering
\includegraphics[width=1\linewidth]{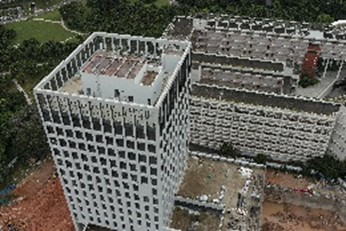}  
\end{minipage}
}
\subfigure{
\begin{minipage}[a]{0.30\linewidth}
\centering
\includegraphics[width=1\linewidth]{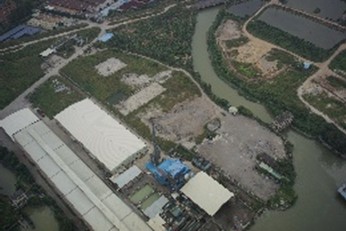}  
\end{minipage}
}
\subfigure{
\begin{minipage}[b]{0.30\linewidth}
\centering
\includegraphics[width=1\linewidth]{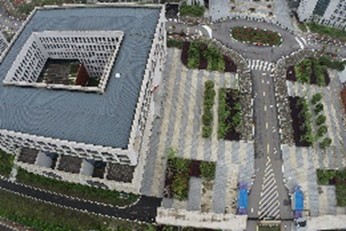} 
\end{minipage}
}
\subfigure{
\begin{minipage}[b]{0.30\linewidth}
\centering
\includegraphics[width=1\linewidth]{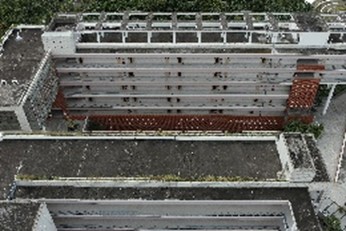} 
\end{minipage}
}
\subfigure{
\begin{minipage}[b]{0.30\linewidth}
\centering
\includegraphics[width=1\linewidth]{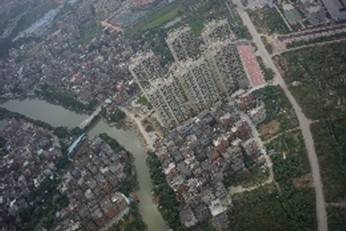} 
\end{minipage}
}
\setcounter{subfigure}{0}
\subfigure[dataset 1]{
\includegraphics[width=0.30\linewidth]{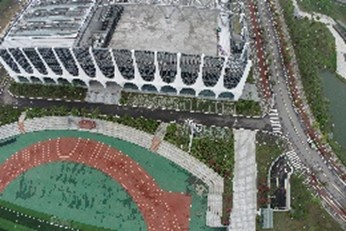}
}
\subfigure[dataset 2]{
\includegraphics[width=0.30\linewidth]{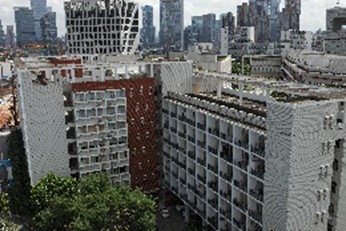}
}
\subfigure[dataset 3]{
\includegraphics[width=0.30\linewidth]{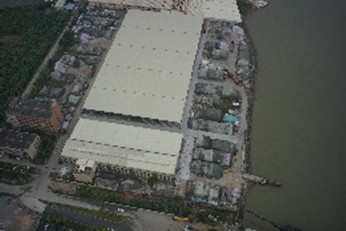}
}
\caption{The illustration of sample images of the three UAV datasets.}
\label{fig:figure 8}
\end{figure}

The data acquisition details of the three test datasets are listed in Table \ref{tab:Table 3}, and the following is a description of each dataset:

\textbf{Dataset 1} is captured using a DJI Phantom 4 RTK UAV equipped with a DJI FC6310R camera over a university campus, as illustrated in Figure \ref{fig:figure 8}(a). The UAV operates at a constant altitude of 80.0 meters above ground level, capturing a total of 3,743 images with a resolution of 5,472 × 3,648 pixels. The ground sampling distance (GSD) is approximately 2.6 centimeters.

\textbf{Dataset 2} is also collected from a university campus, but it specifically focused on a group of complex buildings, as illustrated in Figure \ref{fig:figure 8}(b). Unlike the fixed-altitude data acquisition method, the optimized photogrammetry \citep{li2023optimized} is applied to adjust the shooting direction of the onboard camera according to the geometry structure of the ground target. A DJI Zenmuse P1 camera is used to record a total of 4,030 images, with an image resolution of 8,192 × 5,460 pixels and a GSD of approximately 1.2 centimeters.

\textbf{Dataset 3} covers a large area, including urban buildings and rural bare land, with a long river running through it, as shown in Figure \ref{fig:figure 8}(c). At a flight altitude of 87.0 meters, a classical five-angle oblique photogrammetry system composed of five SONY ILCE 7R cameras is used to capture a total of 21,654 images. For testing, a sub-scene of 4,318 images is selected, with an image resolution of 6,000 × 4,000 pixels and a GSD of approximately 1.2 centimeters.

\begin{table*}[!ht]    
\centering    
\caption{Detailed information about the three UAV datasets}    
\label{tab:Table 3}   
\begin{tabular}{llll} 
\toprule 
        Item name & Dataset 1 & Dataset 2 & Dataset 3  \\        
\midrule 
        UAV type & multi-rotor & multi-rotor & multi-rotor\\
        Flight height (m) & 80.0 & - & 87.0\\
        Camera mode & DJI FC6310R & DJI Zenmuse P1 & SONY ILCE 7R\\
        Number of cameras & 1 & 1 & 5\\
        Focal length (mm) & 24 & 35 & 35\\
        Camera angle (°) & 0 & - & Nadir: 0; oblique: 45/-45\\
        Number of images & 3,743 & 4,030 & 21,654\\
        Image size (pixel) & 5,472×3,648 & 8,192×5,460 & 6,000×4,000\\
        GSD (cm) & 2.6 & 1.2 & 1.2\\   
\bottomrule 
\end{tabular}
\end{table*}

\begin{table*}[!ht]    
\centering   
\caption{Description of the metrics for performance evaluation.}    
\label{tab:Table 4}   
\makebox[1\linewidth]{
\begin{tabular}{l l l} 
\toprule 
        Category & Name & Description  \\        
\midrule 
        \multirow{3}{*}{Match pair retrieval} & Accuracy & The ratio between the number of correct matching pairs \\ & & and the number of retrieved image pairs. \\
        & Efficiency & The total time cost of match pair retrieval. \\
\midrule
         \multirow{3}{*}{3D reconstruction} & Number of registered images & The number of registered images in SfM reconstruction. \\       
         & Number of 3D points & The number of 3D points after sparse reconstruction. \\ 
         & Reprojection error & The RMSE of the bundle adjustment in pixels.\\
\bottomrule 
\end{tabular}
}
\end{table*}

\subsubsection{ Evaluation metrics}
\label{sec4.1.2}

Two categories of evaluation metrics are used to evaluate the image retrieval model. The first category concerns match pair retrieval, including retrieval accuracy and retrieval efficiency. Retrieval accuracy is calculated as shown in Formula \ref{eq19}, where $RP$ represents the set of image pairs retrieved through the match pair retrieval process, $MP$ denotes the set of correct matching pairs retained after feature matching, and $N(*)$ indicates the number of image pairs in the set. Retrieval efficiency is calculated as shown in Formula \ref{eq20}, where $T_{fe}$ represents the time consumed for global feature extraction, and $T_{nns}$ denotes the time required for nearest neighbor searching.$T_{fe}$ and $T_{nns}$ together constitute the total time cost of match pair retrieval.

\begin{equation}    
\text{Accuracy} = \frac{N(MP)}{N(RP)}
\label{eq19}
\end{equation}

\begin{equation}    
{Efficiency} = T_{fe} + T_{nns}
\label{eq20}
\end{equation}

The second category is 3D reconstruction metrics. After performing match pair retrieval and feature matching, the view graph is constructed to guide the parallel SfM described in section \ref{sec2.2} to reconstruct the 3D model of the test scene. The metrics include the completeness and accuracy of the reconstructed model. Completeness is quantified by the number of registered images and the number of reconstructed 3D points, while accuracy is represented by the mean reprojection error. All evaluation metrics are listed in Table \ref{tab:Table 4}.

In the experiments, image pairs with matches greater than 15 are considered correct matching pairs. The retrieval number has a significant impact on the accuracy and efficiency of SfM-based 3D reconstruction. A large retrieval number decreases the efficiency of match pair retrieval and subsequent feature matching, while a small retrieval number may lead to the loss of too many correct matching pairs, potentially resulting in reconstruction failure. Therefore, the retrieval number is fixed to 30 empirically.

\subsection{Experiments setting}
\label{sec4.2}

All the image retrieval models are trained on a Windows computer with 64 GB RAM, four Xeon E5-2680 CPUs, and one 10 GB NVIDIA GeForce RTX 3080 graphics card. The evaluation experiments are executed on a Windows computer with 16 GB memory, one Intel 2.30 GHz i7-12700H CPU, and one 6 GB NVIDIA GeForce RTX 3060 graphics card. The PyTorch framework is employed for experiments, and all the network backbones are initialized with the pre-trained weights from ImageNet. For the implementation of NetVLAD, the clustering centers $c_k$ and assignment function parameters $w_k$ are initialized by utilizing the UAVPairs dataset, while the parameter $p$ of GeM pooling is set to 3. The network training employs the Adam optimizer with an initial learning rate of $l_0=10^{-5}$, which followed an exponential decay schedule $l_i=l_0*exp(-0.1i)$ per epoch, along with a momentum of 0.9 and weight decay of $5\times10^{-4}$. The batch size is 5, and all training images are downsampled to a resolution of 480×320. The training process is limited to 20 epochs, with each epoch consisting of 2,000 iterations. According to \cite{liu2024matchable}, in the nearest neighbor searching of large-scale vectors, the HNSW algorithm can significantly improve the search efficiency while maintaining high accuracy. Therefore, in the test experiments, the deep global features are extracted from images downsampled 5 times, and then the HNSW algorithm implemented by the FAISS library is used to accomplish the nearest neighbor searching.

\subsection{Evaluation in match pair retrieval}
\label{sec4.3}

\subsubsection{The effectiveness of the UAVPairs dataset}
\label{sec4.3.1}

\begin{table*}[!ht]    
\centering   
\caption{Retrieval accuracy comparison of models trained on different datasets (\%)}    
\label{tab:Table 5}   
\begin{tabular}{lllrrr} 
\toprule 
        Model & Backbone & Training dataset & Dataset 1 & Dataset 2 & Dataset 3 \\   
\midrule 
        \multirow{2}{*}{NetVLAD} & \multirow{2}{*}{VGG-16} & Pittsburgh & 81.37 & 87.29 & 72.80\\
        &  &UAVPairs & 85.74 & 88.30 & 73.14 \\
\midrule
        \multirow{2}{*}{GeM} & \multirow{2}{*}{VGG-16} & Flickr & 73.02 & 83.38 & 63.56 \\       
        &  &UAVPairs & 82.09 & 85.52 & 67.02\\
\midrule
        \multirow{2}{*}{MIRorR} & \multirow{2}{*}{ResNet-50} & GL3D & 73.68 & 80.31 & 62.61 \\  &  &UAVPairs & 84.44 & 87.92 & 72.30 \\
\bottomrule 
\end{tabular}
\end{table*}


\begin{figure*}[!h] 
\centering
\subfigure[Retrieval example 1 through NetVLAD trained on the Pittsburgh dataset]{
\includegraphics[width=0.45\linewidth]{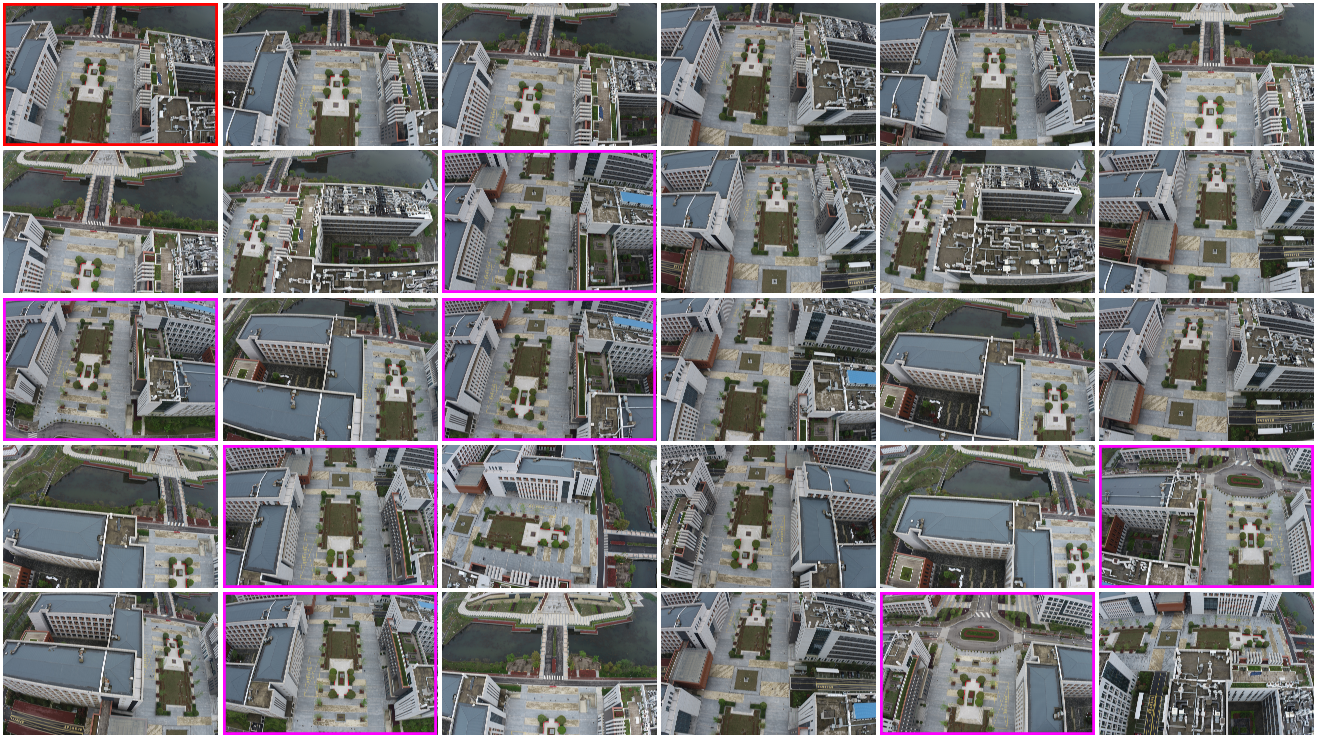}
}
\subfigure[Retrieval example 1 through NetVLAD trained on the UAVPairs dataset]{
\includegraphics[width=0.45\linewidth]{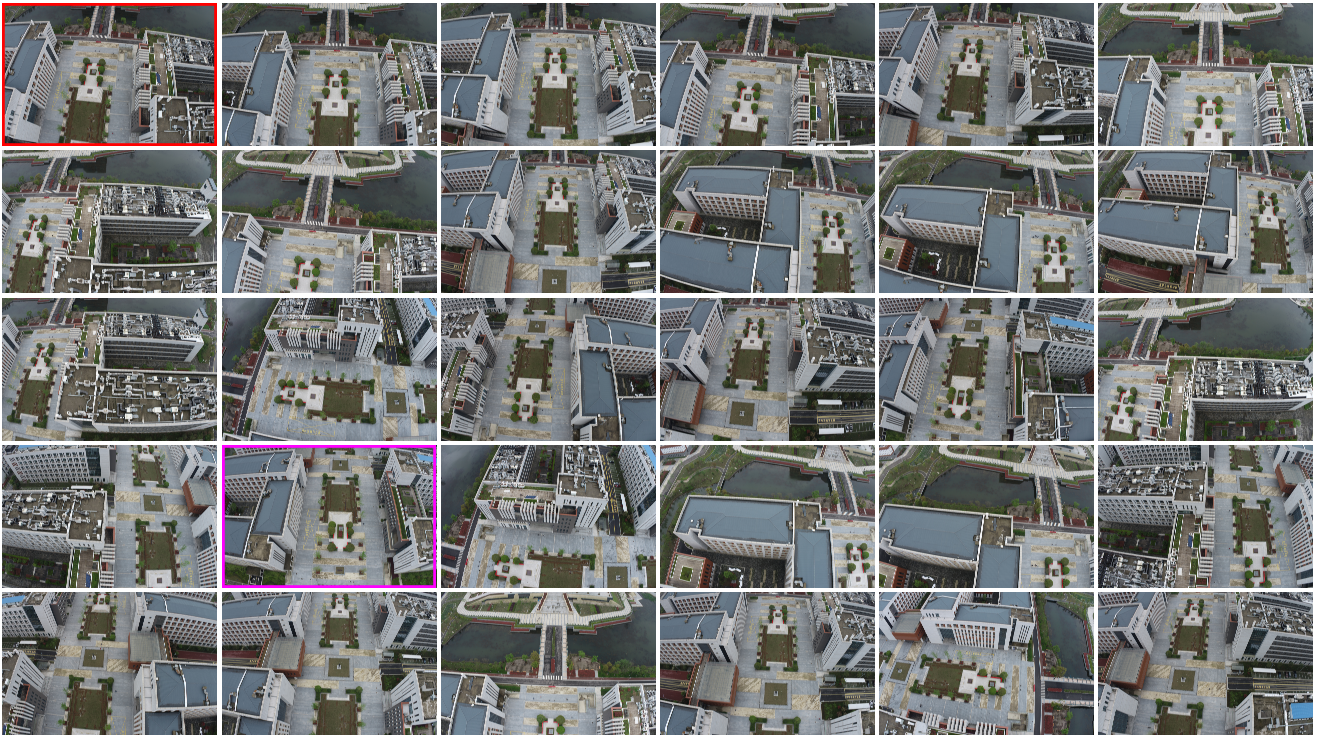}
}
\subfigure[Retrieval example 2 through GeM trained on the Flickr dataset]{
\includegraphics[width=0.45\linewidth]{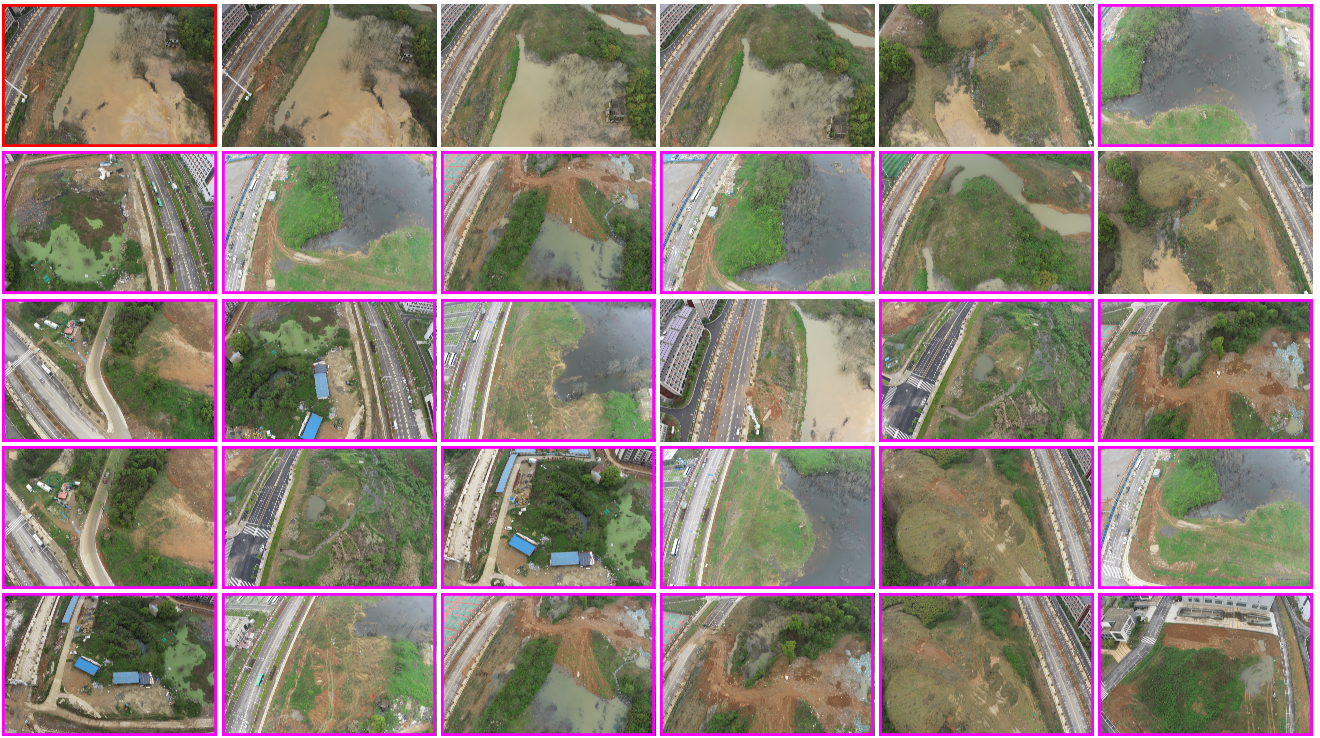}
}
\subfigure[Retrieval example 2 through GeM trained on the UAVPairs dataset]{
\includegraphics[width=0.45\linewidth]{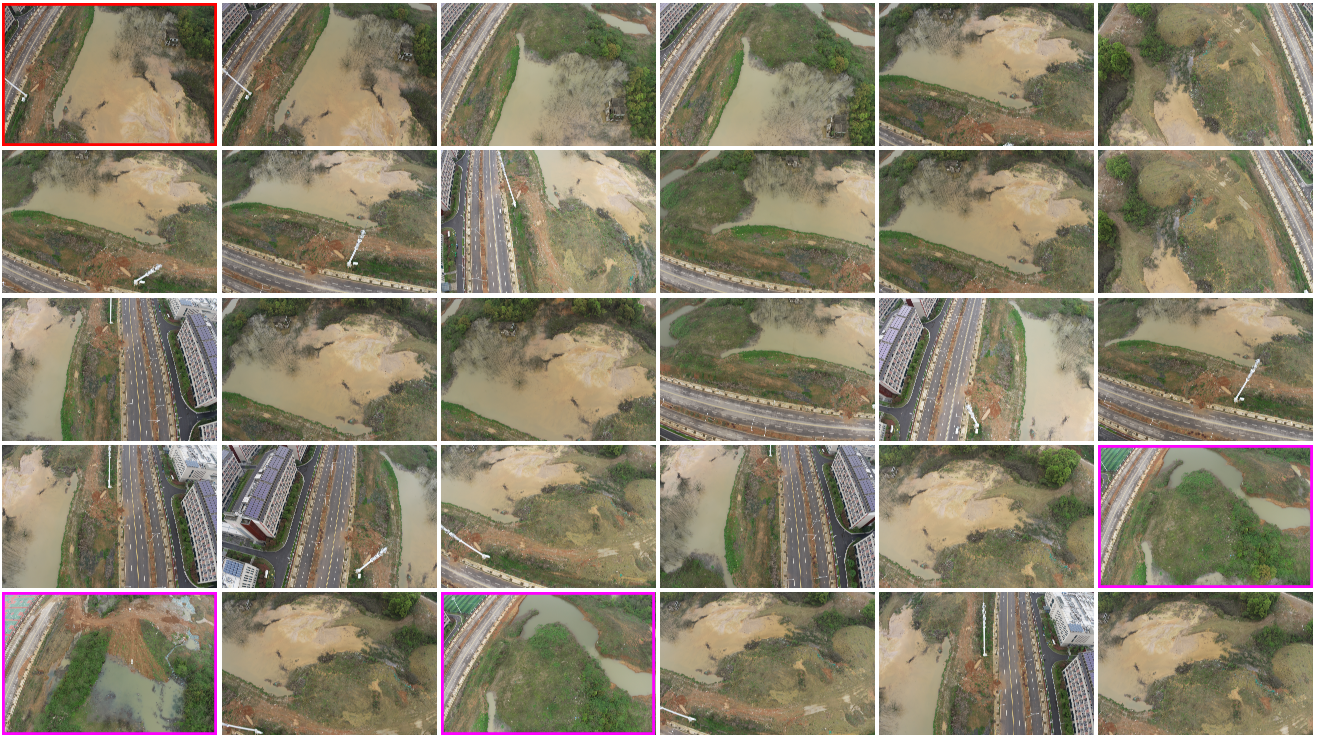}
}
\subfigure[Retrieval example 3 through MIRorR trained on the GL3D dataset]{
\includegraphics[width=0.45\linewidth]{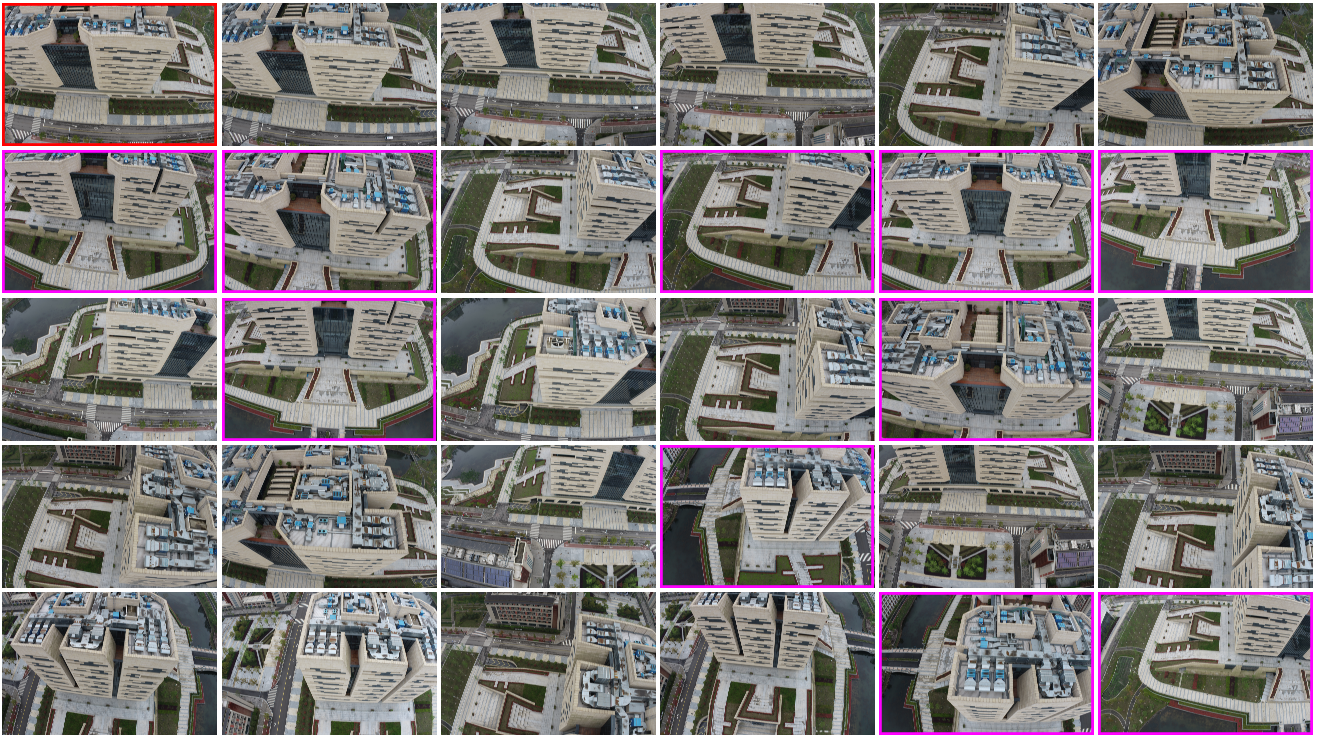}
}
\subfigure[Retrieval example 3 through MIRorR trained on the UAVPairs dataset]{
\includegraphics[width=0.45\linewidth]{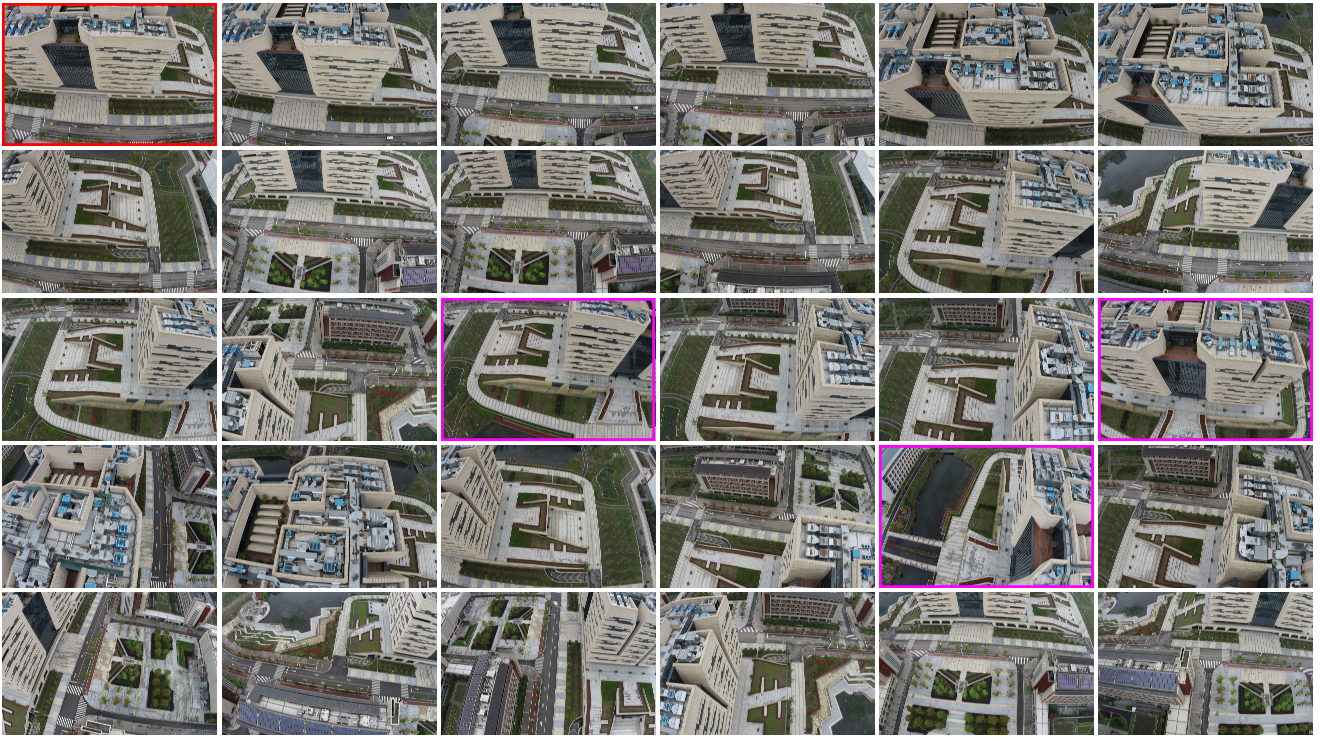}
}
\caption{Retrieval examples of dataset 1 through image retrieval models trained on the UAVPairs dataset and other datasets. (the red box indicates the query image and the purple box indicates the incorrect retrieval image)}
\label{fig:figure 9}
\end{figure*}

The effectiveness of the UAVPairs dataset is verified by comparing the match pair retrieval accuracy of image retrieval models trained with different training datasets. The compared image retrieval models include NetVLAD, GeM, and MIRorR. According to \cite{arandjelovic2016netvlad}, the model with the backbone VGG-16 trained on the Pittsburgh dataset achieved the best test performance, so this model is used as the benchmark model for NetVLAD. The benchmark model for GeM takes the same backbone but is trained on the Flickr dataset. In the experiments of \cite{shen2018matchable}, all models are trained on the GL3D dataset, and the model with the backbone ResNet-50 achieved the highest retrieval accuracy. Therefore, this model is determined as the benchmark model for MIRorR. Then, the three models are trained with the UAVPairs dataset. The training of the models employs triplet loss with the margin parameter $m$ set to 0.1. Before each epoch, global hard negative mining is conducted to generate training samples, where each training sample consists of an anchor image, a positive sample, and two hard negative samples from different scenes. Subsequently, 10,000 sample tuples generated from the UAVPairs dataset are randomly selected for training.

Table \ref{tab:Table 5} shows the retrieval accuracy comparison of models trained on different datasets. Among these, the model NetVLAD trained on the UAVPairs dataset achieves the highest retrieval accuracy. However, the learnable clustering centers and residual assignment parameters of NetVLAD result in higher model complexity, which makes model optimization more difficult. Consequently, its performance improvement is less significant than GeM and MIRorR. Although GeM pooling demonstrates superior feature representation capability over max pooling, the trained GeM model still underperforms MIRorR in retrieval accuracy. This discrepancy is principally attributable to the enhanced feature extraction and generalization capabilities endowed by the ResNet-50 backbone of MIRorR.

The experimental results show that the retrieval accuracy of NetVLAD, GeM, and MIRorR trained on the UAVPairs dataset is improved by an average of 1.9\%, 4.89\%, and 9.35\% on the three test datasets, respectively. Figure \ref{fig:figure 9} presents retrieval examples of dataset 1 through image retrieval models trained on the UAVPairs dataset and other datasets. The retrieval results of the models trained on the UAVPairs dataset few incorrect images, while the model trained on other datasets produces lots of incorrect retrieval images. These results demonstrate that the UAVPairs dataset is more suited for match pair retrieval of large-scale UAV images compared to the existing datasets such as Pittsburgh, Flickr, and GL3D.

\subsubsection{The effectiveness of ranked list loss}
\label{sec4.3.2}

\begin{table*}[!ht]    
\centering   
\caption{Retrieval accuracy comparison of models trained on different datasets (\%)}    
\label{tab:Table 6}   
\begin{tabular}{lllrrr} 
\toprule 
        Model & Loss & Sample Mining & Dataset 1 & Dataset 2 & Dataset 3 \\        
\midrule 
        \multirow{3}{*}{NetVLAD} & Triplet & Global hard negative & 85.74 & 88.30 & 73.14 \\        & Triplet & Batched nontrivial sample & 85.71 & 88.58 & 72.07 \\
        & Ranked List & Batched nontrivial sample & 86.84 & 89.62 & 74.01 \\
\midrule        
        \multirow{3}{*}{GeM} & Triplet & Global hard negative & 82.09 & 85.52 & 67.02 \\          
         & Triplet & Batched nontrivial sample & 80.46 & 85.32 & 65.89 \\
         & Ranked List & Batched nontrivial sample & 82.32 & 85.86 & 68.04 \\
\midrule        
         \multirow{3}{*}{MIRorR} &Triplet & Global hard negative & 84.44 & 87.92 & 72.30 \\
         & Triplet & Batched nontrivial sample & 83.08 & 85.67 & 71.46 \\
         &Ranked List & Batched nontrivial sample & 85.75 & 88.51 & 73.24\\
\bottomrule 
\end{tabular}
\end{table*}

To validate the effectiveness of the ranked list loss, the three models are trained with different loss functions and sample mining methods. For global hard negative mining, the training sample generation follows the identical procedure described in Section \ref{sec4.2}. For batched nontrivial sample mining, we implement the proposed sample generation method as follows: firstly, randomly select $ n$ scenes and sample one image per scene as the anchors; secondly, randomly select $m$ images from overlapping image list of each anchor as the positive samples, which are sorted by geometric similarity; finally, the above process is repeated $t$ times to generate sufficient training samples. To ensure the consistency of batched nontrivial sample mining with global hard negative mining in terms of batch size and iterations, we set $n=5$, $m=3$, and $t=2,000$. The margin parameter $m$ between the positive and negative sample is fixed at 0.1 for both the triplet loss as well as the ranked list loss. In contrast, the margin $\alpha$ between the anchor and positive sample in ranked list loss requires adaptation to different models, and it is set to 1.35, 0.9, and 0.9 for NetVLAD, GeM, and MIRorR, respectively.

\begin{figure}[!ht]
\centering    
\includegraphics[width=1\linewidth]{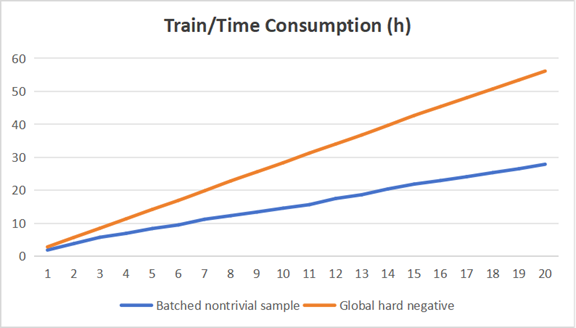}    
\caption{Comparison of training time consumption.}    
\label{fig:figure 10}
\end{figure}

The retrieval accuracy comparison of the models trained with different losses and sample mining methods is presented in Table \ref{tab:Table 6}. Global hard negative mining achieves superior retrieval accuracy compared to batched nontrivial sample mining since it can iteratively select harder triplets with larger loss values. However, it consumes 2 times more training time than batched nontrivial sample mining as shown in Figure \ref{fig:figure 10}, because it requires additional feature extraction of all images in the dataset before training and finding hard-negative samples for each anchor through nearest neighbor searching. The experimental results demonstrate that the three models trained with the proposed ranked list loss and sample mining strategy achieve an average retrieval accuracy improvement of 1.1\%, 0.53\%, and 0.95\%, respectively. When employing the same sample mining method, the ranked list loss achieves average accuracy improvements of 1.37\%, 1.52\%, and 2.43\% over the triplet loss across the three models, respectively. These results confirm that the ranked list loss can effectively enhance the discriminative capability of deep global features.

\subsection{ Evaluation in SfM-based reconstruction}
\label{sec4.4}

\subsubsection{View graph construction}
\label{4.4.1}

\begin{figure*}[!ht]
\centering
\subfigure[Dataset 1: The View graph by NetVLAD]{
\includegraphics[width=0.3\linewidth]{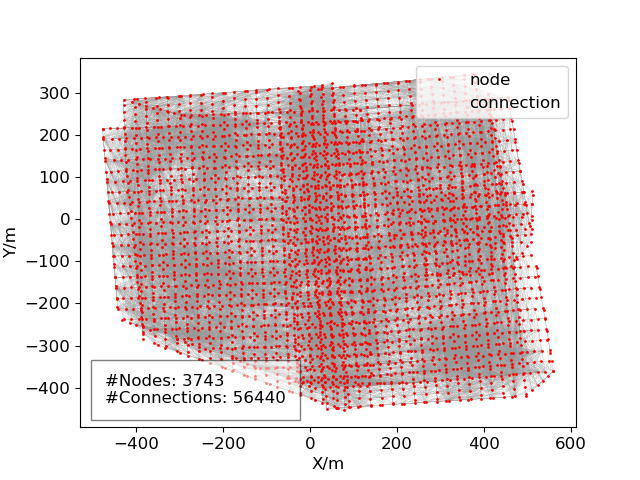}
}
\subfigure[Dataset 2: The View graph by NetVLAD]{
\includegraphics[width=0.3\linewidth]{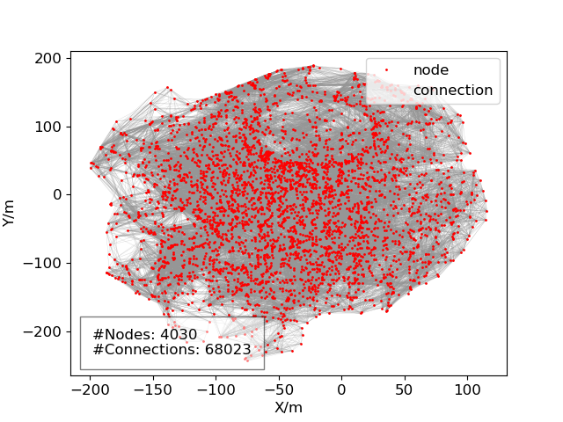}
}
\subfigure[Dataset 3: The View graph by NetVLAD]{
\includegraphics[width=0.3\linewidth]{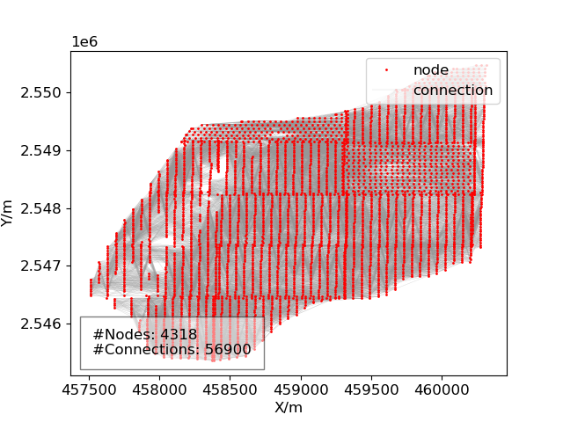}
}
\\
\subfigure[Dataset 1: The View graph by NetVLAD-O]{
\includegraphics[width=0.3\linewidth]{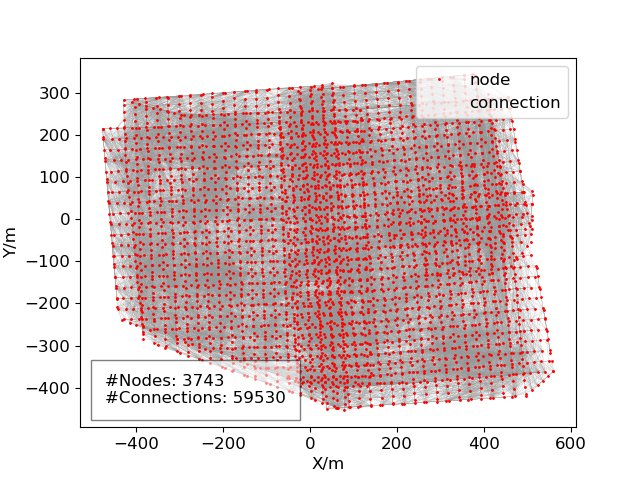}
}
\subfigure[Dataset 2: The View graph by NetVLAD-O]{
\includegraphics[width=0.3\linewidth]{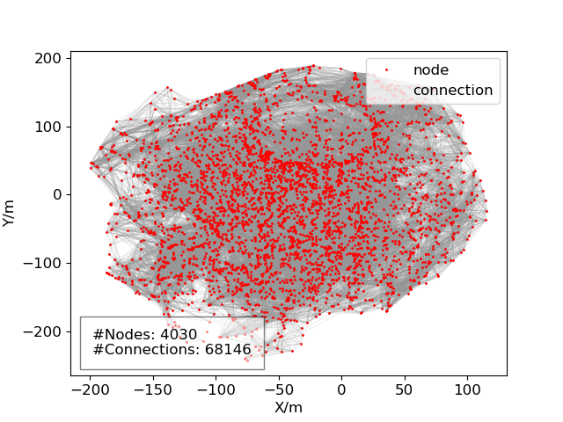}
}
\subfigure[Dataset 3: The View graph by NetVLAD-O]{
\includegraphics[width=0.3\linewidth]{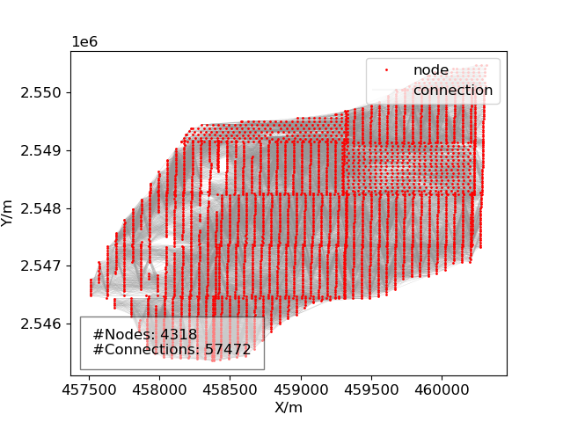}
}
\caption{The view graphs constructed with the match pair retrieval results from NetVLAD and NetVLAD-O.
}
\label{fig:figure 11}
\end{figure*}

The aim of match pair retrieval is to accelerate feature matching in SfM-based 3D reconstruction, and it is essential to compare the results of view graph construction and 3D reconstruction. Since NetVLAD achieves the optimum retrieval accuracy, only the results retrieved by NetVLAD are used in this section. The model trained with the UAVPairs dataset and ranked list loss is termed as NetVLAD-O and the model trained by Arandjelovic et al. is termed as NetVLAD. After obtaining the retrieved image pairs, the parallel SfM reconstruction framework described in Section \ref{sec2.2} is used to construct the view graph and reconstruct the 3D model. Figure \ref{fig:figure 11} shows the view graphs constructed with the match pair retrieval results from NetVLAD and NetVLAD-O. The optimized NetVLAD-O increases 3,090, 123, and 572 connections on the three datasets, respectively, and the constructed view graphs have stronger connectivity, which helps to improve the completeness of reconstruction.

\subsubsection{3D reconstruction}
\label{sec4.4.2}

\begin{figure*}[!h] 
\centering
\subfigure[dataset 1]{
\includegraphics[height=0.2\linewidth]{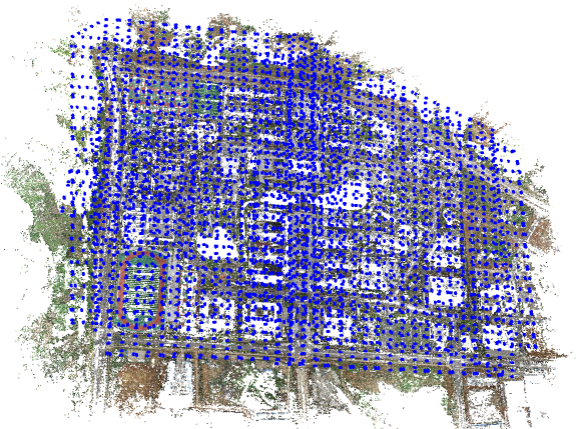}
}
\subfigure[dataset 2]{
\includegraphics[height=0.2\linewidth]{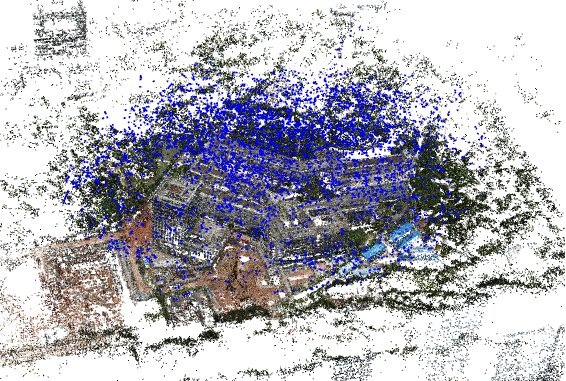}
}
\subfigure[dataset 3]{
\includegraphics[height=0.2\linewidth]{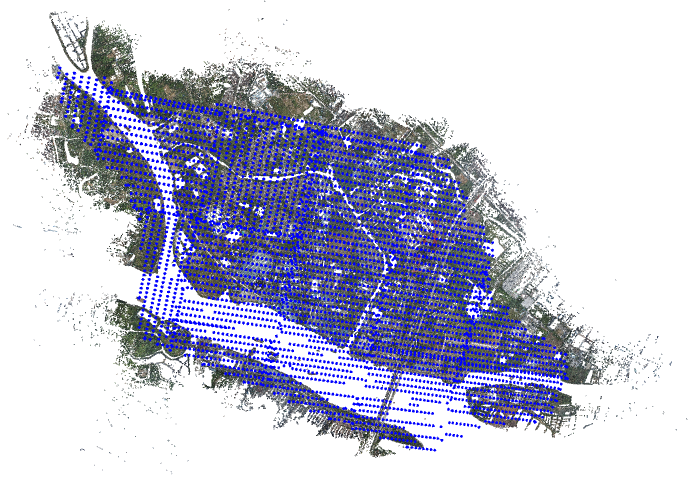}
}
\caption{The reconstructed 3D models of the three datasets. (Registered images are rendered in blue color, and 3D points are colored by image texture.)
}
\label{fig:figure 12}
\end{figure*}

Table \ref{tab:Table 7} presents the statistics of 3D reconstruction implemented with retrieval results from NetVLAD and NetVLAD-O, showing a significant improvement in reconstruction completeness for the three datasets. For datasets 1 and 3, the number of registered images increases by 23 and 31, respectively. The number of reconstructed 3D points increase by 16,880, 9,450, and 19,254 for the three datasets, respectively. As NetVLAD-O reconstructed more 3D points, the reconstruction precision is slightly decreased. However, most of the images in each dataset are successfully registered with a sub-pixel precision of 0.676, 0.800, and 0.756 pixels for the three datasets, respectively. The reconstructed 3D models of the three datasets are shown in Figure \ref{fig:figure 12} for visual analysis. These results demonstrate that the model trained with the UAVPairs dataset and the ranked list loss not only enhances the accuracy of match pair retrieval but also significantly improves the quality of subsequent view graph construction and 3D reconstruction.

\begin{table*}[!ht]
    \centering
    \caption{The statistics of 3D reconstruction implemented with retrieval results from NetVLAD and NetVLAD-O.}    
    \label{tab:Table 7}
    \begin{tabular}{lllrrr}
        \toprule
        Category & Metric & Model & Dataset 1 & Dataset 2 & Dataset 3 \\
        \midrule
        \multirow{4}{*}{Completeness} 
        & Number of & NetVLAD & 3,709/3,743 & 4,029/4,030 & 4,266/4,318 \\
        & registered images & NetVLAD-O & 3,732/3,743 & 4,029/4,030 & 4,297/4,318 \\
        \cmidrule(l){2 - 6} 
        & Number of 3D & NetVLAD & 919,939 & 1,514,529 & 2,069,065\\
        & points & NetVLAD-O & 936,819 & 1,523,979 & 2,088,319 \\
        \midrule
        \multirow{2}{*}{Precision} 
        & Reprojection  & NetVLAD & 0.673 & 0.776 & 0.757 \\ 
        & error (pixel) & NetVLAD-O & 0.676 & 0.780 & 0.758 \\
        \bottomrule
    \end{tabular}
\end{table*}

\subsection{ Compared with other match pair retrieval methods
}
\label{sec4.5}

\begin{figure*}[!ht] 
\centering
\subfigure[Efficiency comparison]{
\includegraphics[width=0.43\linewidth]{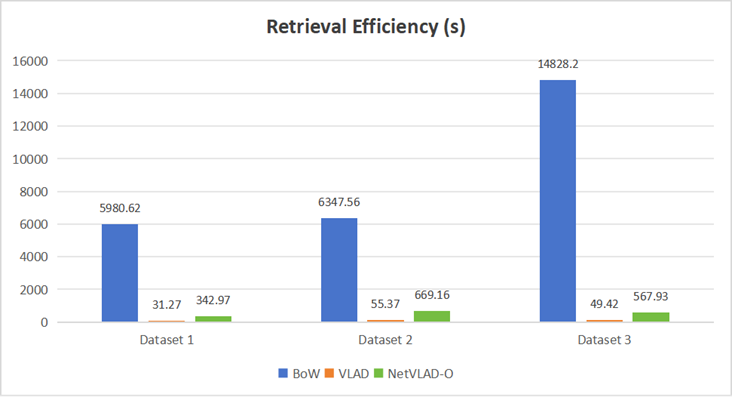}
}
\subfigure[Accuracy comparison]{
\includegraphics[width=0.43\linewidth]{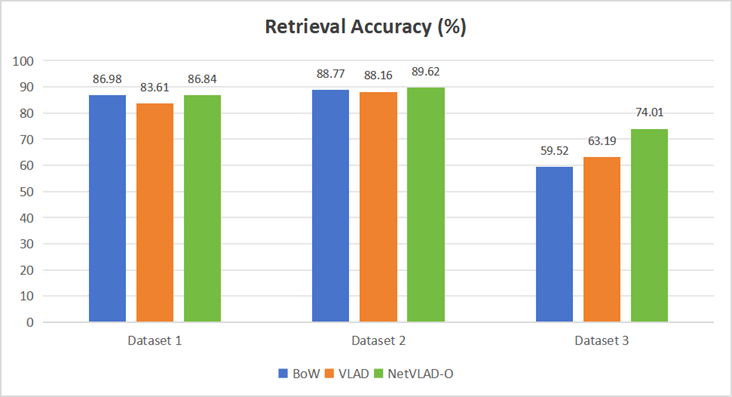}
}
\caption{The efficiency and accuracy comparison of of match pair retrieval methods.}
\label{fig:figure 13}
\end{figure*}

In this section, NetVLAD-O is compared with BoW and VLAD in terms of match pair retrieval and SfM-based 3D reconstruction, which are commonly used in current SfM systems. BoW is implemented with ColMap, where the vocabulary tree is constructed on the Flickr 100k dataset and contains 256K visual words. For VLAD, we adopt the implementation from \cite{jiang2023efficient}, where the codebook size is fixed at 256 words and nearest neighbor searching is performed by HNSW.


\begin{figure*} [!ht]
\centering
\subfigure[Retrieval example 1 through VLAD]{
\includegraphics[width=0.43\linewidth]{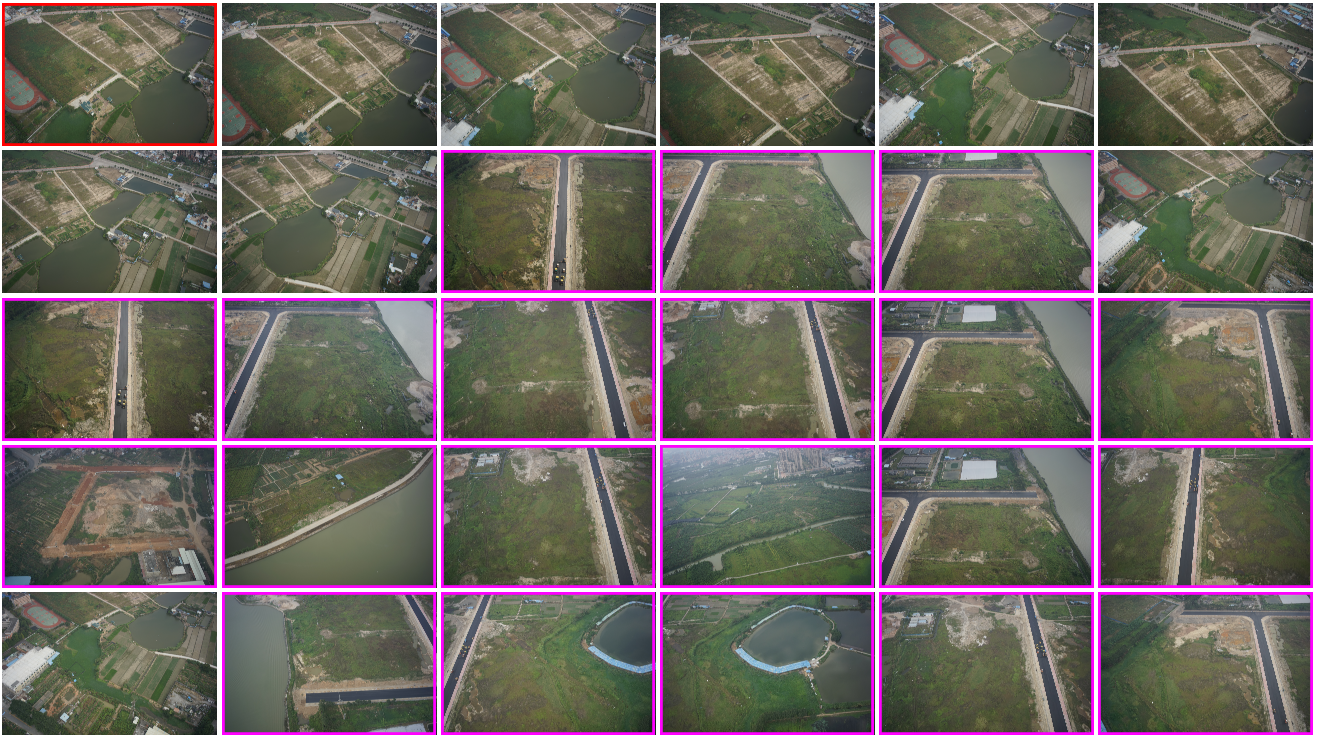}
}
\subfigure[Retrieval example 1 through NetVLAD-O]{
\includegraphics[width=0.43\linewidth]{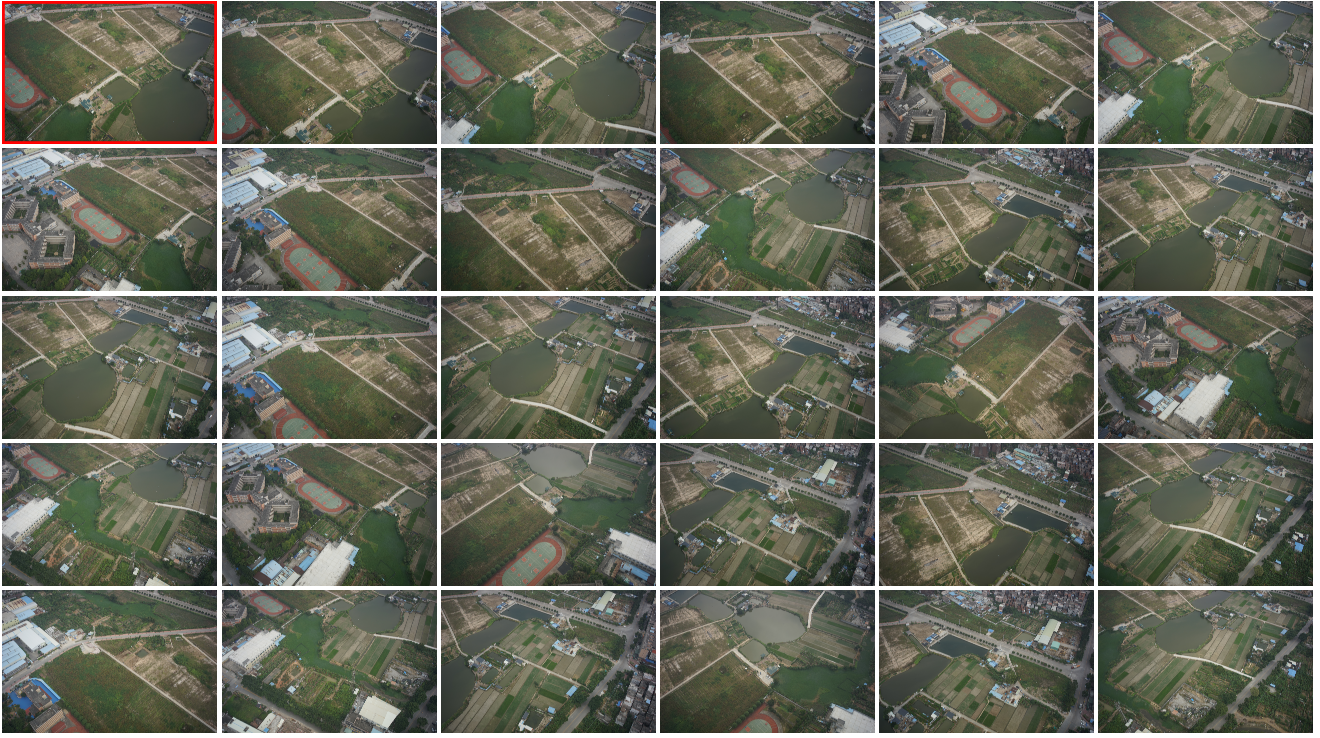}
}
\subfigure[Retrieval example 2 through VLAD]{
\includegraphics[width=0.43\linewidth]{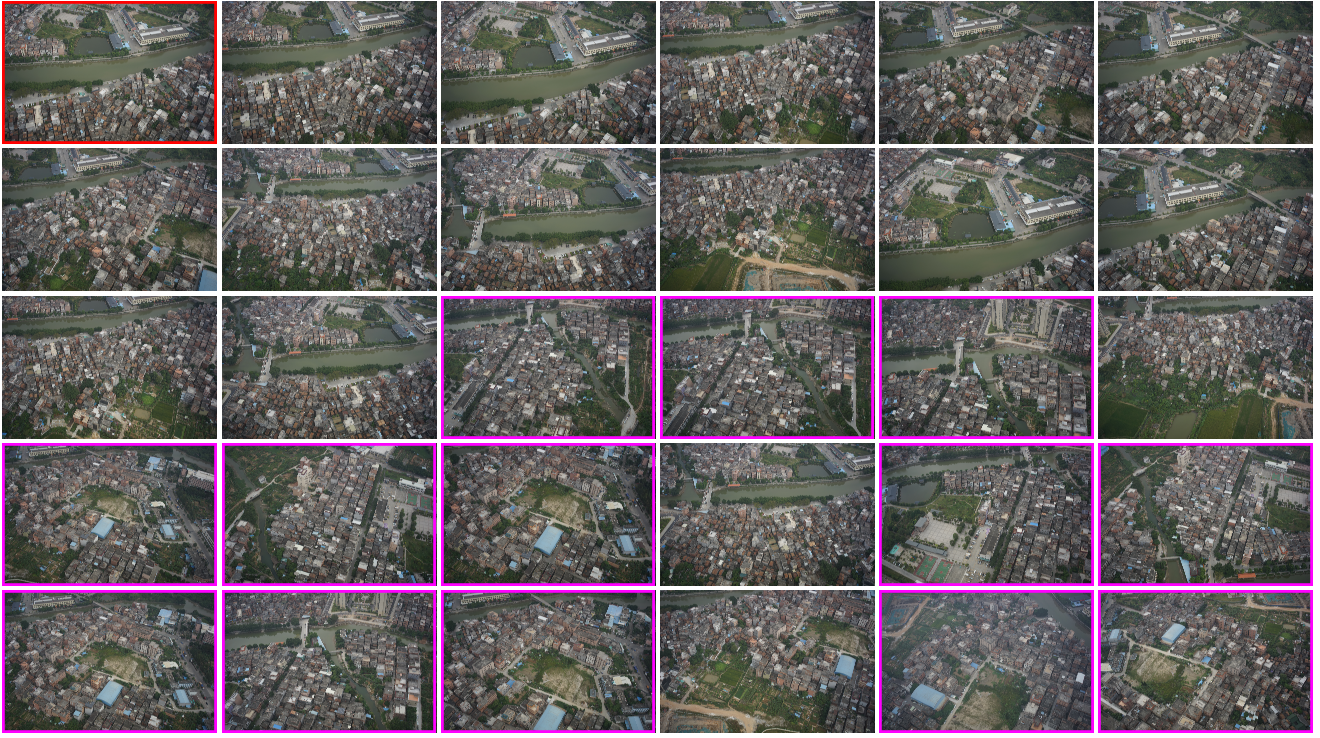}
}
\subfigure[Retrieval example 2 through NetVLAD-O]{
\includegraphics[width=0.43\linewidth]{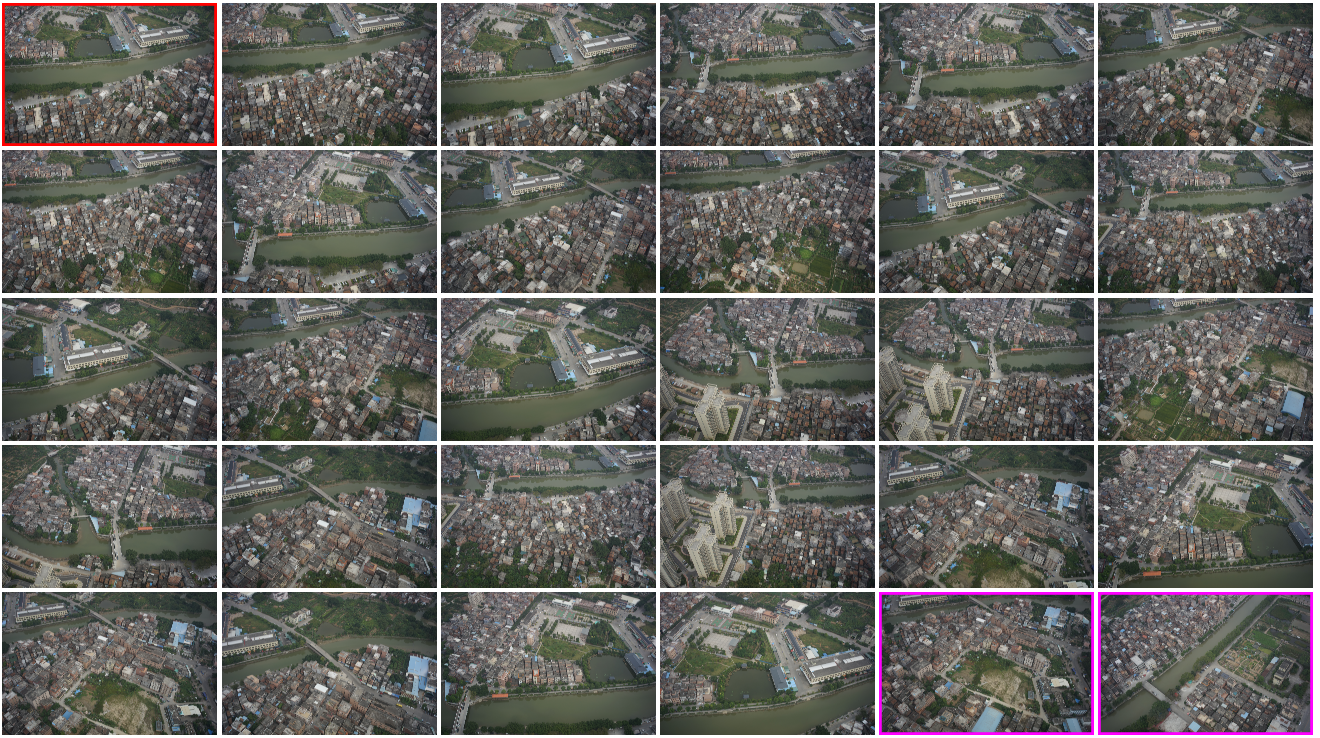}
}
\subfigure[Retrieval example 3 through VLAD]{
\includegraphics[width=0.43\linewidth]{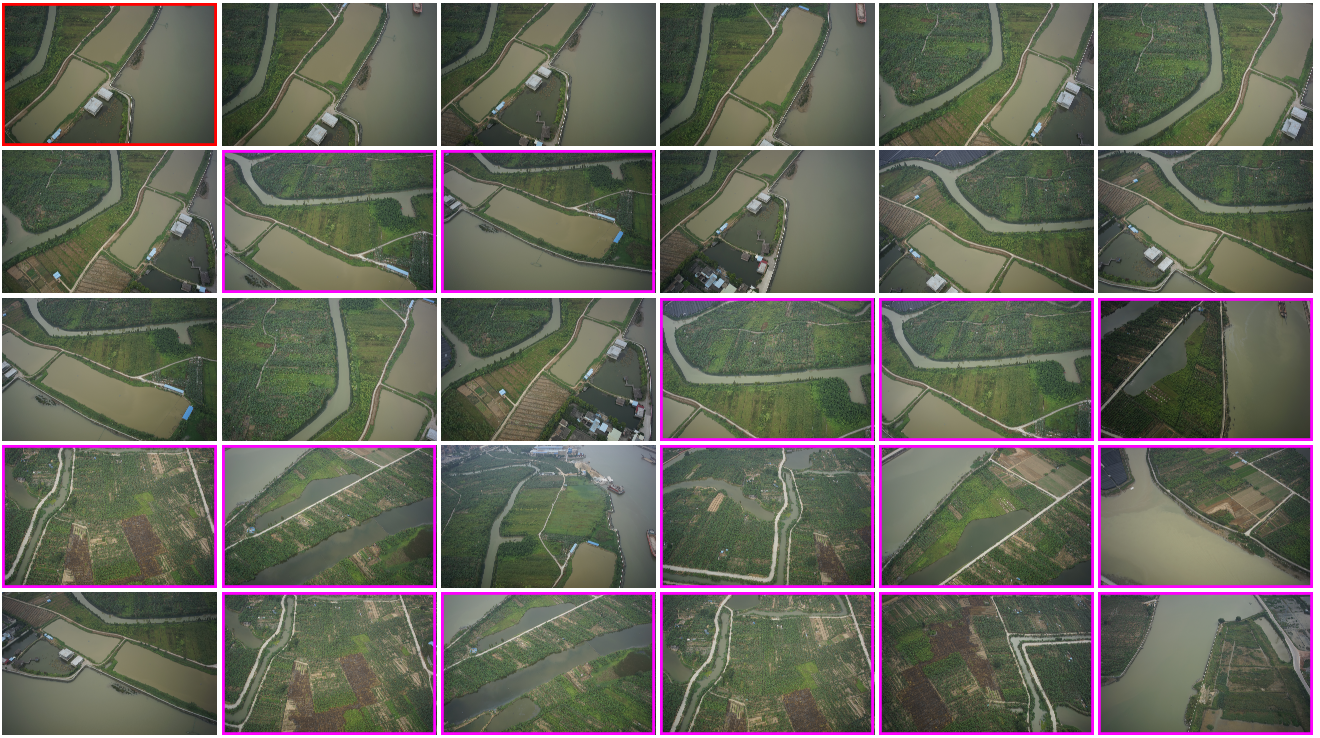}
}
\subfigure[Retrieval example 3 through NetVLAD-O]{
\includegraphics[width=0.43\linewidth]{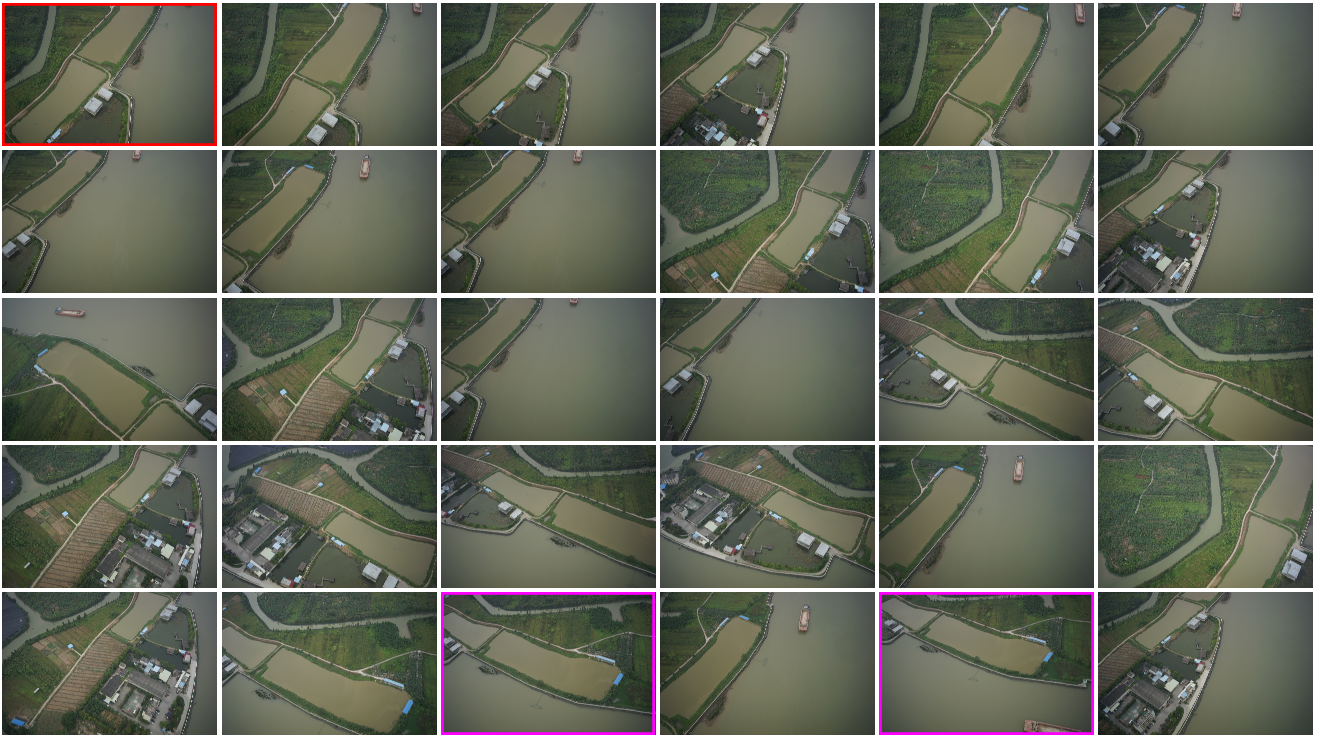}
}
\caption{Retrieval examples of dataset 3 via VLAD and NetVLAD-O. (the red box indicates the query image and the purple box indicates the incorrect retrieval image)}
\label{fig:figure 15}
\end{figure*}

\begin{table*}[!ht]
    \centering
    \caption{The statistics of 3D reconstruction implemented with retrieval results from NetVLAD and NetVLAD-O.}        
    \label{tab:Table 8}
    \begin{tabular}{lllrrr}
        \toprule
       Category & Metric & Model & Dataset 1 & Dataset 2 & Dataset 3 \\
        \midrule
        \multirow{8}{*}{Completeness} 
        & \multirow{4}{*}{Number of registered images} 
        & BoW & 3,716 & 4,029 & 4,267  \\
        &  & VLAD & 3,730 & 4,027 & 4,248 \\
        &  & NetVLAD & 3,709 & 4,029 & 4,266 \\
        &  & NetVLAD-O & 3,732 & 4,029 & 4,297 \\
        \cmidrule(l){2 - 6} 
        & \multirow{4}{*}{Number of 3D points} & BoW & 1,002,275 & 1,514,668 & 2,087,310 \\        
        &  & VLAD & 965,066 & 1,477,146 & 2,080,135 \\      
        &  & NetVLAD & 919,939 & 1,514,529 & 2,069,065 \\   
        &  & NetVLAD-O & 936,819 & 1,523,979 & 2,088,319 \\
        \midrule
        \multirow{4}{*}{Precision} & \multirow{4}{*}{Reprojection error (pixel)}& BoW & 0.703 & 0.809 & 0.758  \\
        &  & VLAD & 0.695 & 0.793 & 0.758 \\
        &  & NetVLAD & 0.673 & 0.776 & 0.757 \\
        & &  NetVLAD-O & 0.676 & 0.780 & 0.758 \\
        \bottomrule
    \end{tabular}
\end{table*}

Figure \ref{fig:figure 13}(a) presents the comparison of retrieval efficiency of match pair retrieval methods. The results demonstrate that NetVLAD-O achieves significantly higher retrieval efficiency than BoW, delivering a speedup ratio ranging from 9 to 26. Within the SfM framework, local feature extraction is an imperative processing stage, and as such its computational overhead is excluded from the computation of retrieval efficiency. For hand-crafted global features, the retrieval efficiency consists only of the time consumption for feature aggregation and nearest neighbor searching, whereas deep global features incur additional computational overhead from their dedicated feature extraction process. This explains why NetVLAD-O demonstrates lower retrieval efficiency compared with VLAD.

Figure \ref{fig:figure 13}(b) presents the comparison of retrieval accuracy of match pair retrieval methods. On dataset 1, NetVLAD-O exhibits marginally lower retrieval accuracy than BoW, whereas it achieves the highest retrieval accuracy on datasets 2 and 3. The proposed method not only accelerates match pair retrieval significantly but also outperforms the conventional BoW approach in terms of retrieval accuracy. Moreover, the experimental results on dataset 3 demonstrate that the deep global feature outperforms the handcrafted global feature in repetitively textured scenes and weakly textured scenes, further validating the superiority of NetVLAD-O. As illustrated in examples 1 and 2 of Figure {\ref{fig:figure 15}}, VLAD struggles with scenes containing extensive repetitive textures since the relied local feature SIFT only focuses on local image patches. Furthermore, due to the absence of keypoints in weakly textured regions, VLAD primarily aggregates local features from rich textured regions of images, resulting in poor retrieval accuracy for weakly textured scenes, as shown in Example 3 of Figure {\ref{fig:figure 15}}. In contrast, NetVLAD-O converges global contexts of images and accounts for two-view geometric relationships between image pairs during training, enabling robust performance in both repetitively textured scenes and weakly textured scenes. Table \ref{tab:Table 8} presents the statistics of 3D reconstruction implemented with retrieval results from different match pair retrieval methods. The experimental results demonstrate that NetVLAD-O exhibits higher reconstruction completeness compared to BoW and VLAD on dataset 2 and dataset 3. However, for Dataset 1, although NetVLAD-O registers more images more but not as many reconstructed 3D points as BoW and VLAD, mainly due to the fact that local features are taken into account in the BoW-based or VLAD-based image retrieval while NetVLAD-O does not.

\section{Conclusion}
\label{sec5}

In this study, we have proposed a benchmark and training pipeline for match pair retrieval of large-scale UAV images. Three main contributions have been made to address existing challenges. On the one hand, the UAVPairs dataset is constructed, utilizing SfM-based 3D reconstruction to define geometric similarity for annotating image pairs, ensuring that the image pairs used for training are genuinely matchable. On the other hand, to improve training efficiency and model discrimination, a batched nontrivial sample mining strategy is proposed to decrease mining cost, and the ranked list loss is designed to leverage global similarity structures, overcoming the limitation of other pair-based losses. The experimental results demonstrate that models trained using the UAVPairs dataset and the ranked list loss showed improved retrieval accuracy and enhanced SfM reconstruction quality compared to the baseline models and traditional methods. For match pair retrieval of large-scale UAV images, especially in challenging scenes, the image retrieval model trained with the proposed benchmark and training pipeline can be an effective solution.

\section*{Acknowledgment}
This research was funded by the National Natural Science Foundation of China (Grant No. 42371442, 42301514), and the Hubei Provincial Natural Science Foundation of China (Grant No. 2023AFB568).

\bibliographystyle{elsarticle-harv} 
\bibliography{cas-refs}

\begin{thebibliography}{44}
\expandafter\ifx\csname natexlab\endcsname\relax\def\natexlab#1{#1}\fi
\providecommand{\url}[1]{\texttt{#1}}
\providecommand{\href}[2]{#2}
\providecommand{\path}[1]{#1}
\providecommand{\DOIprefix}{doi:}
\providecommand{\ArXivprefix}{arXiv:}
\providecommand{\URLprefix}{URL: }
\providecommand{\Pubmedprefix}{pmid:}
\providecommand{\doi}[1]{\href{http://dx.doi.org/#1}{\path{#1}}}
\providecommand{\Pubmed}[1]{\href{pmid:#1}{\path{#1}}}
\providecommand{\bibinfo}[2]{#2}
\ifx\xfnm\relax \def\xfnm[#1]{\unskip,\space#1}\fi
\bibitem[{Arandjelovic et~al.(2016)Arandjelovic, Gronat, Torii, Pajdla and
  Sivic}]{arandjelovic2016netvlad}
\bibinfo{author}{Arandjelovic, R.}, \bibinfo{author}{Gronat, P.},
  \bibinfo{author}{Torii, A.}, \bibinfo{author}{Pajdla, T.},
  \bibinfo{author}{Sivic, J.}, \bibinfo{year}{2016}.
\newblock \bibinfo{title}{Netvlad: Cnn architecture for weakly supervised place
  recognition}, in: \bibinfo{booktitle}{Proceedings of the IEEE conference on
  computer vision and pattern recognition}, pp. \bibinfo{pages}{5297--5307}.
\bibitem[{Deng et~al.(2019)Deng, Guo, Xue and Zafeiriou}]{8953658}
\bibinfo{author}{Deng, J.}, \bibinfo{author}{Guo, J.}, \bibinfo{author}{Xue,
  N.}, \bibinfo{author}{Zafeiriou, S.}, \bibinfo{year}{2019}.
\newblock \bibinfo{title}{Arcface: Additive angular margin loss for deep face
  recognition}, in: \bibinfo{booktitle}{2019 IEEE/CVF Conference on Computer
  Vision and Pattern Recognition (CVPR)}, pp. \bibinfo{pages}{4685--4694}.
\bibitem[{Detone et~al.(2018)Detone, Malisiewicz and
  Rabinovich}]{2018SuperPoint}
\bibinfo{author}{Detone, D.}, \bibinfo{author}{Malisiewicz, T.},
  \bibinfo{author}{Rabinovich, A.}, \bibinfo{year}{2018}.
\newblock \bibinfo{title}{Superpoint: Self-supervised interest point detection
  and description}, in: \bibinfo{booktitle}{2018 IEEE/CVF Conference on
  Computer Vision and Pattern Recognition Workshops (CVPRW)}.
\bibitem[{Dusmanu et~al.(2019)Dusmanu, Rocco, Pajdla, Pollefeys, Sivic, Torii
  and Sattler}]{8953622}
\bibinfo{author}{Dusmanu, M.}, \bibinfo{author}{Rocco, I.},
  \bibinfo{author}{Pajdla, T.}, \bibinfo{author}{Pollefeys, M.},
  \bibinfo{author}{Sivic, J.}, \bibinfo{author}{Torii, A.},
  \bibinfo{author}{Sattler, T.}, \bibinfo{year}{2019}.
\newblock \bibinfo{title}{D2-net: A trainable cnn for joint description and
  detection of local features}, in: \bibinfo{booktitle}{2019 IEEE/CVF
  Conference on Computer Vision and Pattern Recognition (CVPR)}, pp.
  \bibinfo{pages}{8084--8093}.
\bibitem[{Hartmann et~al.(2016)Hartmann, Havlena and Schindler}]{2016Recent}
\bibinfo{author}{Hartmann, W.}, \bibinfo{author}{Havlena, M.},
  \bibinfo{author}{Schindler, K.}, \bibinfo{year}{2016}.
\newblock \bibinfo{title}{Recent developments in large-scale tie-point
  matching}.
\newblock \bibinfo{journal}{Isprs Journal of Photogrammetry and Remote Sensing}
  \bibinfo{volume}{115}, \bibinfo{pages}{47--62}.
\bibitem[{Herbert et~al.(2008)Herbert, Bay, , , Andreas, Ess, , , Tinne,
  Tuytelaars, ,  and and}]{Herbert2008Speeded}
\bibinfo{author}{Herbert}, \bibinfo{author}{Bay}, , ,
  \bibinfo{author}{Andreas}, \bibinfo{author}{Ess}, , ,
  \bibinfo{author}{Tinne}, \bibinfo{author}{Tuytelaars}, , ,
  \bibinfo{author}{and, L.}, \bibinfo{year}{2008}.
\newblock \bibinfo{title}{Speeded-up robust features (surf)}.
\newblock \bibinfo{journal}{Computer Vision and Image Understanding} .
\bibitem[{Hou et~al.(2023)Hou, Xia, Zhang, Feng, Zhan and Wang}]{HOU2023103162}
\bibinfo{author}{Hou, Q.}, \bibinfo{author}{Xia, R.}, \bibinfo{author}{Zhang,
  J.}, \bibinfo{author}{Feng, Y.}, \bibinfo{author}{Zhan, Z.},
  \bibinfo{author}{Wang, X.}, \bibinfo{year}{2023}.
\newblock \bibinfo{title}{Learning visual overlapping image pairs for sfm via
  cnn fine-tuning with photogrammetric geometry information}.
\newblock \bibinfo{journal}{International Journal of Applied Earth Observation
  and Geoinformation} \bibinfo{volume}{116}, \bibinfo{pages}{103162}.
\bibitem[{Jegou et~al.(2008)Jegou, Douze and Schmid}]{jegou2008hamming}
\bibinfo{author}{Jegou, H.}, \bibinfo{author}{Douze, M.},
  \bibinfo{author}{Schmid, C.}, \bibinfo{year}{2008}.
\newblock \bibinfo{title}{Hamming embedding and weak geometric consistency for
  large scale image search}, in: \bibinfo{booktitle}{Computer Vision--ECCV
  2008: 10th European Conference on Computer Vision, Marseille, France, October
  12-18, 2008, Proceedings, Part I 10}, \bibinfo{organization}{Springer}. pp.
  \bibinfo{pages}{304--317}.
\bibitem[{J{\'e}gou et~al.(2010)J{\'e}gou, Douze, Schmid and
  P{\'e}rez}]{jegou2010aggregating}
\bibinfo{author}{J{\'e}gou, H.}, \bibinfo{author}{Douze, M.},
  \bibinfo{author}{Schmid, C.}, \bibinfo{author}{P{\'e}rez, P.},
  \bibinfo{year}{2010}.
\newblock \bibinfo{title}{Aggregating local descriptors into a compact image
  representation}, in: \bibinfo{booktitle}{2010 IEEE computer society
  conference on computer vision and pattern recognition},
  \bibinfo{organization}{IEEE}. pp. \bibinfo{pages}{3304--3311}.
\bibitem[{Jiang et~al.(2021)Jiang, Jiang and Wang}]{jiang2021unmanned}
\bibinfo{author}{Jiang, S.}, \bibinfo{author}{Jiang, W.},
  \bibinfo{author}{Wang, L.}, \bibinfo{year}{2021}.
\newblock \bibinfo{title}{Unmanned aerial vehicle-based photogrammetric 3d
  mapping: A survey of techniques, applications, and challenges}.
\newblock \bibinfo{journal}{IEEE Geoscience and Remote Sensing Magazine}
  \bibinfo{volume}{10}, \bibinfo{pages}{135--171}.
\bibitem[{Jiang et~al.(2022)Jiang, Li, Jiang and Chen}]{jiang2022parallel}
\bibinfo{author}{Jiang, S.}, \bibinfo{author}{Li, Q.}, \bibinfo{author}{Jiang,
  W.}, \bibinfo{author}{Chen, W.}, \bibinfo{year}{2022}.
\newblock \bibinfo{title}{Parallel structure from motion for uav images via
  weighted connected dominating set}.
\newblock \bibinfo{journal}{IEEE Transactions on Geoscience and Remote Sensing}
  \bibinfo{volume}{60}, \bibinfo{pages}{1--13}.
\bibitem[{Jiang et~al.(2023)Jiang, Ma, Liu, Li, Jiang, Guo, Li and
  Wang}]{jiang2023efficient}
\bibinfo{author}{Jiang, S.}, \bibinfo{author}{Ma, Y.}, \bibinfo{author}{Liu,
  J.}, \bibinfo{author}{Li, Q.}, \bibinfo{author}{Jiang, W.},
  \bibinfo{author}{Guo, B.}, \bibinfo{author}{Li, L.}, \bibinfo{author}{Wang,
  L.}, \bibinfo{year}{2023}.
\newblock \bibinfo{title}{Efficient match pair retrieval for large-scale uav
  images via graph indexed global descriptor}.
\newblock \bibinfo{journal}{IEEE Journal of Selected Topics in Applied Earth
  Observations and Remote Sensing} \bibinfo{volume}{16},
  \bibinfo{pages}{9874--9887}.
\bibitem[{Li et~al.(2023)Li, Huang, Yu and Jiang}]{li2023optimized}
\bibinfo{author}{Li, Q.}, \bibinfo{author}{Huang, H.}, \bibinfo{author}{Yu,
  W.}, \bibinfo{author}{Jiang, S.}, \bibinfo{year}{2023}.
\newblock \bibinfo{title}{Optimized views photogrammetry: Precision analysis
  and a large-scale case study in qingdao}.
\newblock \bibinfo{journal}{IEEE Journal of Selected Topics in Applied Earth
  Observations and Remote Sensing} \bibinfo{volume}{16},
  \bibinfo{pages}{1144--1159}.
\bibitem[{Liu et~al.(2024)Liu, Ma, Jiang, Wang, Li and
  Jiang}]{liu2024matchable}
\bibinfo{author}{Liu, J.}, \bibinfo{author}{Ma, Y.}, \bibinfo{author}{Jiang,
  S.}, \bibinfo{author}{Wang, L.}, \bibinfo{author}{Li, Q.},
  \bibinfo{author}{Jiang, W.}, \bibinfo{year}{2024}.
\newblock \bibinfo{title}{Matchable image retrieval for large-scale uav images:
  an evaluation of sfm-based reconstruction}.
\newblock \bibinfo{journal}{International Journal of Remote Sensing}
  \bibinfo{volume}{45}, \bibinfo{pages}{692--718}.
\bibitem[{Lowe(2004)}]{lowe2004distinctive}
\bibinfo{author}{Lowe, D.G.}, \bibinfo{year}{2004}.
\newblock \bibinfo{title}{Distinctive image features from scale-invariant
  keypoints}.
\newblock \bibinfo{journal}{International journal of computer vision}
  \bibinfo{volume}{60}, \bibinfo{pages}{91--110}.
\bibitem[{Luo et~al.(2018)Luo, Shen, Zhou, Zhu, Zhang, Yao, Fang and
  Quan}]{luo2018geodesc}
\bibinfo{author}{Luo, Z.}, \bibinfo{author}{Shen, T.}, \bibinfo{author}{Zhou,
  L.}, \bibinfo{author}{Zhu, S.}, \bibinfo{author}{Zhang, R.},
  \bibinfo{author}{Yao, Y.}, \bibinfo{author}{Fang, T.}, \bibinfo{author}{Quan,
  L.}, \bibinfo{year}{2018}.
\newblock \bibinfo{title}{Geodesc: Learning local descriptors by integrating
  geometry constraints}, in: \bibinfo{booktitle}{Proceedings of the European
  conference on computer vision (ECCV)}, pp. \bibinfo{pages}{168--183}.
\bibitem[{Luo et~al.(2020)Luo, Zhou, Bai, Chen, Zhang, Yao, Li, Fang and
  Quan}]{luo2020aslfeat}
\bibinfo{author}{Luo, Z.}, \bibinfo{author}{Zhou, L.}, \bibinfo{author}{Bai,
  X.}, \bibinfo{author}{Chen, H.}, \bibinfo{author}{Zhang, J.},
  \bibinfo{author}{Yao, Y.}, \bibinfo{author}{Li, S.}, \bibinfo{author}{Fang,
  T.}, \bibinfo{author}{Quan, L.}, \bibinfo{year}{2020}.
\newblock \bibinfo{title}{Aslfeat: Learning local features of accurate shape
  and localization}, in: \bibinfo{booktitle}{Proceedings of the IEEE/CVF
  conference on computer vision and pattern recognition}, pp.
  \bibinfo{pages}{6589--6598}.
\bibitem[{Mishchuk et~al.(2017)Mishchuk, Mishkin, Radenovic and
  Matas}]{mishchuk2017working}
\bibinfo{author}{Mishchuk, A.}, \bibinfo{author}{Mishkin, D.},
  \bibinfo{author}{Radenovic, F.}, \bibinfo{author}{Matas, J.},
  \bibinfo{year}{2017}.
\newblock \bibinfo{title}{Working hard to know your neighbor's margins: Local
  descriptor learning loss}.
\newblock \bibinfo{journal}{Advances in neural information processing systems}
  \bibinfo{volume}{30}.
\bibitem[{Movshovitz-Attias et~al.(2017)Movshovitz-Attias, Toshev, Leung, Ioffe
  and Singh}]{movshovitz2017no}
\bibinfo{author}{Movshovitz-Attias, Y.}, \bibinfo{author}{Toshev, A.},
  \bibinfo{author}{Leung, T.K.}, \bibinfo{author}{Ioffe, S.},
  \bibinfo{author}{Singh, S.}, \bibinfo{year}{2017}.
\newblock \bibinfo{title}{No fuss distance metric learning using proxies}, in:
  \bibinfo{booktitle}{Proceedings of the IEEE international conference on
  computer vision}, pp. \bibinfo{pages}{360--368}.
\bibitem[{Ng et~al.(2020)Ng, Balntas, Tian and Mikolajczyk}]{ng2020solar}
\bibinfo{author}{Ng, T.}, \bibinfo{author}{Balntas, V.}, \bibinfo{author}{Tian,
  Y.}, \bibinfo{author}{Mikolajczyk, K.}, \bibinfo{year}{2020}.
\newblock \bibinfo{title}{Solar: second-order loss and attention for image
  retrieval}, in: \bibinfo{booktitle}{Computer Vision--ECCV 2020: 16th European
  Conference, Glasgow, UK, August 23--28, 2020, Proceedings, Part XXV 16},
  \bibinfo{organization}{Springer}. pp. \bibinfo{pages}{253--270}.
\bibitem[{Nister and Stewenius(2006)}]{nister2006scalable}
\bibinfo{author}{Nister, D.}, \bibinfo{author}{Stewenius, H.},
  \bibinfo{year}{2006}.
\newblock \bibinfo{title}{Scalable recognition with a vocabulary tree}, in:
  \bibinfo{booktitle}{2006 IEEE Computer Society Conference on Computer Vision
  and Pattern Recognition (CVPR'06)}, \bibinfo{organization}{Ieee}. pp.
  \bibinfo{pages}{2161--2168}.
\bibitem[{Noh et~al.(2017)Noh, Araujo, Sim, Weyand and Han}]{noh2017large}
\bibinfo{author}{Noh, H.}, \bibinfo{author}{Araujo, A.}, \bibinfo{author}{Sim,
  J.}, \bibinfo{author}{Weyand, T.}, \bibinfo{author}{Han, B.},
  \bibinfo{year}{2017}.
\newblock \bibinfo{title}{Large-scale image retrieval with attentive deep local
  features}, in: \bibinfo{booktitle}{Proceedings of the IEEE international
  conference on computer vision}, pp. \bibinfo{pages}{3456--3465}.
\bibitem[{Perronnin et~al.(2010)Perronnin, Liu, S{\'a}nchez and
  Poirier}]{perronnin2010large}
\bibinfo{author}{Perronnin, F.}, \bibinfo{author}{Liu, Y.},
  \bibinfo{author}{S{\'a}nchez, J.}, \bibinfo{author}{Poirier, H.},
  \bibinfo{year}{2010}.
\newblock \bibinfo{title}{Large-scale image retrieval with compressed fisher
  vectors}, in: \bibinfo{booktitle}{2010 IEEE computer society conference on
  computer vision and pattern recognition}, \bibinfo{organization}{IEEE}. pp.
  \bibinfo{pages}{3384--3391}.
\bibitem[{Philbin et~al.(2007)Philbin, Chum, Isard, Sivic and
  Zisserman}]{philbin2007object}
\bibinfo{author}{Philbin, J.}, \bibinfo{author}{Chum, O.},
  \bibinfo{author}{Isard, M.}, \bibinfo{author}{Sivic, J.},
  \bibinfo{author}{Zisserman, A.}, \bibinfo{year}{2007}.
\newblock \bibinfo{title}{Object retrieval with large vocabularies and fast
  spatial matching}, in: \bibinfo{booktitle}{2007 IEEE conference on computer
  vision and pattern recognition}, \bibinfo{organization}{IEEE}. pp.
  \bibinfo{pages}{1--8}.
\bibitem[{Philbin et~al.(2008)Philbin, Chum, Isard, Sivic and
  Zisserman}]{philbin2008lost}
\bibinfo{author}{Philbin, J.}, \bibinfo{author}{Chum, O.},
  \bibinfo{author}{Isard, M.}, \bibinfo{author}{Sivic, J.},
  \bibinfo{author}{Zisserman, A.}, \bibinfo{year}{2008}.
\newblock \bibinfo{title}{Lost in quantization: Improving particular object
  retrieval in large scale image databases}, in: \bibinfo{booktitle}{2008 IEEE
  conference on computer vision and pattern recognition},
  \bibinfo{organization}{IEEE}. pp. \bibinfo{pages}{1--8}.
\bibitem[{Radenovi{\'c} et~al.(2018)Radenovi{\'c}, Tolias and
  Chum}]{radenovic2018fine}
\bibinfo{author}{Radenovi{\'c}, F.}, \bibinfo{author}{Tolias, G.},
  \bibinfo{author}{Chum, O.}, \bibinfo{year}{2018}.
\newblock \bibinfo{title}{Fine-tuning cnn image retrieval with no human
  annotation}.
\newblock \bibinfo{journal}{IEEE transactions on pattern analysis and machine
  intelligence} \bibinfo{volume}{41}, \bibinfo{pages}{1655--1668}.
\bibitem[{Revaud et~al.(2019)Revaud, De~Souza, Humenberger and
  Weinzaepfel}]{revaud2019r2d2}
\bibinfo{author}{Revaud, J.}, \bibinfo{author}{De~Souza, C.},
  \bibinfo{author}{Humenberger, M.}, \bibinfo{author}{Weinzaepfel, P.},
  \bibinfo{year}{2019}.
\newblock \bibinfo{title}{R2d2: Reliable and repeatable detector and
  descriptor}.
\newblock \bibinfo{journal}{Advances in neural information processing systems}
  \bibinfo{volume}{32}.
\bibitem[{ROOPAK et~al.(1993)ROOPAK, SHAH, EDUARD, SCKINGER, JAMES, W., BENTZ,
  ISABELLE, GUYON and CLIFF}]{ROOPAK1993SIGNATURE}
\bibinfo{author}{ROOPAK}, \bibinfo{author}{SHAH}, \bibinfo{author}{EDUARD},
  \bibinfo{author}{SCKINGER}, \bibinfo{author}{JAMES}, \bibinfo{author}{W.},
  \bibinfo{author}{BENTZ}, \bibinfo{author}{ISABELLE}, \bibinfo{author}{GUYON},
  \bibinfo{author}{CLIFF}, \bibinfo{year}{1993}.
\newblock \bibinfo{title}{Signature verification using a "siamese" time delay
  neural network}.
\newblock \bibinfo{journal}{International Journal of Pattern Recognition and
  Artificial Intelligence} \bibinfo{volume}{07}, \bibinfo{pages}{669--669}.
\bibitem[{Rublee et~al.(2011)Rublee, Rabaud, Konolige and
  Bradski}]{rublee2011orb}
\bibinfo{author}{Rublee, E.}, \bibinfo{author}{Rabaud, V.},
  \bibinfo{author}{Konolige, K.}, \bibinfo{author}{Bradski, G.},
  \bibinfo{year}{2011}.
\newblock \bibinfo{title}{Orb: An efficient alternative to sift or surf}, in:
  \bibinfo{booktitle}{2011 International conference on computer vision},
  \bibinfo{organization}{Ieee}. pp. \bibinfo{pages}{2564--2571}.
\bibitem[{Schonberger and Frahm(2016)}]{schonberger2016structure}
\bibinfo{author}{Schonberger, J.L.}, \bibinfo{author}{Frahm, J.M.},
  \bibinfo{year}{2016}.
\newblock \bibinfo{title}{Structure-from-motion revisited}, in:
  \bibinfo{booktitle}{Proceedings of the IEEE conference on computer vision and
  pattern recognition}, pp. \bibinfo{pages}{4104--4113}.
\bibitem[{Schroff et~al.(2015)Schroff, Kalenichenko and
  Philbin}]{schroff2015facenet}
\bibinfo{author}{Schroff, F.}, \bibinfo{author}{Kalenichenko, D.},
  \bibinfo{author}{Philbin, J.}, \bibinfo{year}{2015}.
\newblock \bibinfo{title}{Facenet: A unified embedding for face recognition and
  clustering}, in: \bibinfo{booktitle}{Proceedings of the IEEE conference on
  computer vision and pattern recognition}, pp. \bibinfo{pages}{815--823}.
\bibitem[{Shen et~al.(2018)Shen, Luo, Zhou, Zhang, Zhu, Fang and
  Quan}]{shen2018matchable}
\bibinfo{author}{Shen, T.}, \bibinfo{author}{Luo, Z.}, \bibinfo{author}{Zhou,
  L.}, \bibinfo{author}{Zhang, R.}, \bibinfo{author}{Zhu, S.},
  \bibinfo{author}{Fang, T.}, \bibinfo{author}{Quan, L.}, \bibinfo{year}{2018}.
\newblock \bibinfo{title}{Matchable image retrieval by learning from surface
  reconstruction}, in: \bibinfo{booktitle}{Asian conference on computer
  vision}, \bibinfo{organization}{Springer}. pp. \bibinfo{pages}{415--431}.
\bibitem[{Shi and Malik(2000)}]{shi2000normalized}
\bibinfo{author}{Shi, J.}, \bibinfo{author}{Malik, J.}, \bibinfo{year}{2000}.
\newblock \bibinfo{title}{Normalized cuts and image segmentation}.
\newblock \bibinfo{journal}{IEEE Transactions on pattern analysis and machine
  intelligence} \bibinfo{volume}{22}, \bibinfo{pages}{888--905}.
\bibitem[{Sivic and Zisserman(2003)}]{sivic2003video}
\bibinfo{author}{Sivic}, \bibinfo{author}{Zisserman}, \bibinfo{year}{2003}.
\newblock \bibinfo{title}{Video google: A text retrieval approach to object
  matching in videos}, in: \bibinfo{booktitle}{Proceedings ninth IEEE
  international conference on computer vision}, \bibinfo{organization}{IEEE}.
  pp. \bibinfo{pages}{1470--1477}.
\bibitem[{Song et~al.(2022a)Song, Han and Avrithis}]{song2022all}
\bibinfo{author}{Song, C.H.}, \bibinfo{author}{Han, H.J.},
  \bibinfo{author}{Avrithis, Y.}, \bibinfo{year}{2022}a.
\newblock \bibinfo{title}{All the attention you need: Global-local,
  spatial-channel attention for image retrieval}, in:
  \bibinfo{booktitle}{Proceedings of the IEEE/CVF winter conference on
  applications of computer vision}, pp. \bibinfo{pages}{2754--2763}.
\bibitem[{Song et~al.(2022b)Song, Zhu, Yang and He}]{song2022dalg}
\bibinfo{author}{Song, Y.}, \bibinfo{author}{Zhu, R.}, \bibinfo{author}{Yang,
  M.}, \bibinfo{author}{He, D.}, \bibinfo{year}{2022}b.
\newblock \bibinfo{title}{Dalg: Deep attentive local and global modeling for
  image retrieval}.
\newblock \bibinfo{journal}{arXiv preprint arXiv:2207.00287} .
\bibitem[{Tian et~al.(2017)Tian, Fan and Wu}]{tian2017l2}
\bibinfo{author}{Tian, Y.}, \bibinfo{author}{Fan, B.}, \bibinfo{author}{Wu,
  F.}, \bibinfo{year}{2017}.
\newblock \bibinfo{title}{L2-net: Deep learning of discriminative patch
  descriptor in euclidean space}, in: \bibinfo{booktitle}{Proceedings of the
  IEEE conference on computer vision and pattern recognition}, pp.
  \bibinfo{pages}{661--669}.
\bibitem[{Wang et~al.(2018)Wang, Wang, Zhou, Ji, Gong, Zhou, Li and
  Liu}]{wang2018cosface}
\bibinfo{author}{Wang, H.}, \bibinfo{author}{Wang, Y.}, \bibinfo{author}{Zhou,
  Z.}, \bibinfo{author}{Ji, X.}, \bibinfo{author}{Gong, D.},
  \bibinfo{author}{Zhou, J.}, \bibinfo{author}{Li, Z.}, \bibinfo{author}{Liu,
  W.}, \bibinfo{year}{2018}.
\newblock \bibinfo{title}{Cosface: Large margin cosine loss for deep face
  recognition}, in: \bibinfo{booktitle}{Proceedings of the IEEE conference on
  computer vision and pattern recognition}, pp. \bibinfo{pages}{5265--5274}.
\bibitem[{Wang and Jiang(2015)}]{wang2015instre}
\bibinfo{author}{Wang, S.}, \bibinfo{author}{Jiang, S.}, \bibinfo{year}{2015}.
\newblock \bibinfo{title}{Instre: a new benchmark for instance-level object
  retrieval and recognition}.
\newblock \bibinfo{journal}{ACM Transactions on Multimedia Computing,
  Communications, and Applications (TOMM)} \bibinfo{volume}{11},
  \bibinfo{pages}{1--21}.
\bibitem[{Wang et~al.(2019)Wang, Rottensteiner and Heipke}]{wang2019structure}
\bibinfo{author}{Wang, X.}, \bibinfo{author}{Rottensteiner, F.},
  \bibinfo{author}{Heipke, C.}, \bibinfo{year}{2019}.
\newblock \bibinfo{title}{Structure from motion for ordered and unordered image
  sets based on random kd forests and global pose estimation}.
\newblock \bibinfo{journal}{ISPRS Journal of Photogrammetry and Remote Sensing}
  \bibinfo{volume}{147}, \bibinfo{pages}{19--41}.
\bibitem[{Weyand et~al.(2020)Weyand, Araujo, Cao and Sim}]{weyand2020google}
\bibinfo{author}{Weyand, T.}, \bibinfo{author}{Araujo, A.},
  \bibinfo{author}{Cao, B.}, \bibinfo{author}{Sim, J.}, \bibinfo{year}{2020}.
\newblock \bibinfo{title}{Google landmarks dataset v2-a large-scale benchmark
  for instance-level recognition and retrieval}, in:
  \bibinfo{booktitle}{Proceedings of the IEEE/CVF conference on computer vision
  and pattern recognition}, pp. \bibinfo{pages}{2575--2584}.
\bibitem[{Yandex and Lempitsky(2015)}]{7410507}
\bibinfo{author}{Yandex, A.B.}, \bibinfo{author}{Lempitsky, V.},
  \bibinfo{year}{2015}.
\newblock \bibinfo{title}{Aggregating local deep features for image retrieval},
  in: \bibinfo{booktitle}{2015 IEEE International Conference on Computer Vision
  (ICCV)}, pp. \bibinfo{pages}{1269--1277}.
\bibitem[{Yang et~al.(2021)Yang, He, Fan, Shi, Xue, Li, Ding and
  Huang}]{yang2021dolg}
\bibinfo{author}{Yang, M.}, \bibinfo{author}{He, D.}, \bibinfo{author}{Fan,
  M.}, \bibinfo{author}{Shi, B.}, \bibinfo{author}{Xue, X.},
  \bibinfo{author}{Li, F.}, \bibinfo{author}{Ding, E.}, \bibinfo{author}{Huang,
  J.}, \bibinfo{year}{2021}.
\newblock \bibinfo{title}{Dolg: Single-stage image retrieval with deep
  orthogonal fusion of local and global features}, in:
  \bibinfo{booktitle}{Proceedings of the IEEE/CVF International conference on
  Computer Vision}, pp. \bibinfo{pages}{11772--11781}.
\bibitem[{Zhang et~al.(2024)Zhang, Huang, Li, Wang and Zhou}]{zhang2024legged}
\bibinfo{author}{Zhang, X.}, \bibinfo{author}{Huang, Z.}, \bibinfo{author}{Li,
  Q.}, \bibinfo{author}{Wang, R.}, \bibinfo{author}{Zhou, B.},
  \bibinfo{year}{2024}.
\newblock \bibinfo{title}{Legged robot-aided 3d tunnel mapping via residual
  compensation and anomaly detection}.
\newblock \bibinfo{journal}{ISPRS Journal of Photogrammetry and Remote Sensing}
  \bibinfo{volume}{214}, \bibinfo{pages}{33--47}.

\end{thebibliography}





\end{document}